\documentstyle[12pt,chicagob,url]{article}

\newtheorem{THEOREM}{Theorem}[section]
\newenvironment{theorem}{\begin{THEOREM} \hspace{-.85em} {\bf :} }%
                        {\end{THEOREM}}
\newtheorem{LEMMA}[THEOREM]{Lemma}
\newenvironment{lemma}{\begin{LEMMA} \hspace{-.85em} {\bf :} }%
                      {\end{LEMMA}}
\newtheorem{COROLLARY}[THEOREM]{Corollary}
\newenvironment{corollary}{\begin{COROLLARY} \hspace{-.85em} {\bf :} }%
                          {\end{COROLLARY}}
\newtheorem{PROPOSITION}[THEOREM]{Proposition}
\newenvironment{proposition}{\begin{PROPOSITION} \hspace{-.85em} {\bf :} }%
                            {\end{PROPOSITION}}
\newtheorem{DEFINITION}[THEOREM]{Definition}
\newenvironment{definition}{\begin{DEFINITION} \hspace{-.85em} {\bf :} \rm}%
                            {\end{DEFINITION}}
\newtheorem{CLAIM}[THEOREM]{Claim}
\newenvironment{claim}{\begin{CLAIM} \hspace{-.85em} {\bf :} \rm}%
                            {\end{CLAIM}}
\newtheorem{EXAMPLE}[THEOREM]{Example}
\newenvironment{example}{\begin{EXAMPLE} \hspace{-.85em} {\bf :} \rm}%
                            {\end{EXAMPLE}}
\newtheorem{REMARK}[THEOREM]{Remark}
\newenvironment{remark}{\begin{REMARK} \hspace{-.85em} {\bf :} \rm}%
                            {\end{REMARK}}

\newcommand{\thm}{\begin{theorem}}
\newcommand{\lem}{\begin{lemma}}
\newcommand{\pro}{\begin{proposition}}
\newcommand{\dfn}{\begin{definition}}
\newcommand{\rem}{\begin{remark}}
\newcommand{\xam}{\begin{example}}
\newcommand{\cor}{\begin{corollary}}
\newcommand{\prf}{\noindent{\bf Proof:} }
\newcommand{\ethm}{\end{theorem}}
\newcommand{\elem}{\end{lemma}}
\newcommand{\epro}{\end{proposition}}
\newcommand{\edfn}{\bbox\end{definition}}
\newcommand{\erem}{\bbox\end{remark}}
\newcommand{\exam}{\bbox\end{example}}
\newcommand{\ecor}{\end{corollary}}
\newcommand{\eprf}{\bbox\vspace{0.1in}}
\newcommand{\beqn}{\begin{equation}}
\newcommand{\eeqn}{\end{equation}}

\newcommand{\bbox}{\vrule height7pt width4pt depth1pt}

\newcommand{\clm}{\begin{claim}}
\newcommand{\eclm}{\end{claim}}

\newcommand{\sat}{\models}

\newcommand{\rimp}{\Rightarrow}

\newcommand{\dimp}{\Leftrightarrow}

\newcommand{\union}{\cup}
\newcommand{\inter}{\cap}

\renewcommand{\phi}{\varphi}

\newcommand{\C}{{\cal C}}

\newcommand{\K}{{\cal K}}
\newcommand{\M}{{\cal M}}

\renewcommand{\P}{{\cal P}}

\newcommand{\U}{{\cal U}}

\newcommand{\eg}{e.g.,~}
\newcommand{\ie}{i.e.,~}

\newcommand{\respc}{resp.,\ }

\newcommand{\ol}{\setlength{\itemsep}{0pt}\begin{enumerate}}
\newcommand{\eol}{\end{enumerate}\setlength{\itemsep}{-\parsep}}
\newcommand{\ul}{\setlength{\itemsep}{0pt}\begin{itemize}}
\newcommand{\dl}{\setlength{\itemsep}{0pt}\begin{description}}
\newcommand{\edl}{\end{description}\setlength{\itemsep}{-\parsep}}
\newcommand{\eul}{\end{itemize}\setlength{\itemsep}{-\parsep}}
\newtheorem{fthm}{Theorem}
\newtheorem{flem}[fthm]{Lemma}
\newtheorem{fcor}[fthm]{Corollary}

\newcommand{\false}{\mbox{{\it false}}}

\newcommand{\commentout}[1]{}

\newcommand{\bi}{\begin{itemize}}
\newcommand{\ei}{\end{itemize}}
\newcommand{\be}{\begin{enumerate}}
\newcommand{\ee}{\end{enumerate}}

\newcommand{\denselist}{\itemsep 0pt\partopsep 0pt}

\commentout{stuff for Baltzer:

\newcommand{\sat}{\models}

\newcommand{\rimp}{\Rightarrow}
\newcommand{\dimp}{\Leftrightarrow}
\newcommand{\union}{\cup}
\newcommand{\inter}{\cap}
\renewcommand{\phi}{\varphi}

\newcommand{\C}{{\cal C}}

\newcommand{\K}{{\cal K}}
\newcommand{\M}{{\cal M}}

\renewcommand{\P}{{\cal P}}

\newcommand{\U}{{\cal U}}

\newcommand{\eg}{e.g.,~}
\newcommand{\ie}{i.e.,~}

\newcommand{\respc}{resp.,\ }

\newcommand{\ol}{\setlength{\itemsep}{0pt}\begin{enumerate}}
\newcommand{\eol}{\end{enumerate}\setlength{\itemsep}{-\parsep}}
\newcommand{\ul}{\setlength{\itemsep}{0pt}\begin{itemize}}
\newcommand{\dl}{\setlength{\itemsep}{0pt}\begin{description}}
\newcommand{\edl}{\end{description}\setlength{\itemsep}{-\parsep}}
\newcommand{\eul}{\end{itemize}\setlength{\itemsep}{-\parsep}}

\newcommand{\false}{\mbox{{\it false}}}

\newcommand{\commentout}[1]{}

\newcommand{\bi}{\begin{itemize}}
\newcommand{\ei}{\end{itemize}}
\newcommand{\be}{\begin{enumerate}}
\newcommand{\ee}{\end{enumerate}}

\newcommand{\denselist}{\itemsep 0pt\partopsep 0pt}

\newcommand{\thmm}{\begin{thm}}
\newcommand{\ethm}{\end{thm}}
\newcommand{\lemm}{\begin{lem}}

\newcommand{\elem}{\end{lem}}
\newcommand{\prom}{\begin{prop}}
\newcommand{\epro}{\end{prop}}
\newcommand{\prf}{\begin{pf}}
\newcommand{\eprf}{\end{pf}}
\newcommand{\xam}{\begin{exmp}}
\newcommand{\exam}{\end{exmp}}
\newcommand{\corm}{\begin{cor}}
\newcommand{\ecor}{\end{cor}}
\newcommand{\citeyear}{\cite}
}%

\newcommand{\best}{f_\preceq}
\newcommand{\RCond}{\stackrel{e}{\rightarrow}}
\newcommand{\Cond}{\rightarrow}

\renewcommand{\S}{{\cal S}}
\newcommand{\intension}[1]{[\![ #1 ]\!]}
\renewcommand{\Omega}{W}
\renewcommand{\omega}{w}
\newcommand{\world}{w}
\newcommand{\Ktt}{{\sf K}}
\newcommand{\ftt}{{\sf f}}
\newcommand{\LPK}{{\cal L}^K}
\newcommand{\LPC}{{\cal L}^C}
\newcommand{\thmm}{\thm}
\newcommand{\lemm}{\lem}
\newcommand{\prom}{\pro}
\newcommand{\corm}{\cor}

\begin{document}

\begin{titlepage}
\title{Set-Theoretic Completeness for Epistemic and Conditional Logic%
\thanks{This paper is almost identical to one that will appear in {\em
Annals of Mathematics and Artificial Intelligence} in 1999.  The work was
supported in part by NSF under grant IRI-96-25901 and by the Air Force
Office of  Scientific Research under grant F49620-96-1-0323.  A
preliminary version of this work appeared in the {\em Fifth 
International Symposium on Articial Intelligence and Mathematics}, 1998.}}
\author{Joseph Y.\ Halpern\\
Computer Science Dept.\\
Cornell University \\
   Ithaca, NY 14853\\
   halpern@cs.cornell.edu\\
   http://www.cs.cornell.edu/home/halpern}
\date{ }
\maketitle
\thispagestyle{empty}

\begin{abstract}
The standard approach to logic in the literature in philosophy and
mathematics, which has also been adopted in computer science, is to
define a language (the {\em syntax}),
an appropriate class of models together with an
interpretation of formulas in the language  (the {\em semantics}),
a collection of axioms and rules of inference characterizing reasoning
(the {\em proof theory}), and then relate the proof theory to the
semantics via soundness and completeness results.
Here we consider an approach that is more common in the economics
literature, which works purely at the semantic, set-theoretic level.
We provide set-theoretic completeness results for a number of epistemic
and conditional logics, and contrast the expressive power of the
syntactic and set-theoretic approaches.
\end{abstract}

\end{titlepage}

\section{Introduction}
The standard approach to logic in the literature in philosophy and
mathematics, which has also been adopted in computer science, is to
define a language (the {\em syntax}),
an appropriate class of models together with an
interpretation of formulas in the language (the
{\em semantics}),
a collection of axioms and rules of inference characterizing reasoning
(the {\em proof theory}), and then relate the proof theory to the
semantics via soundness and completeness results.

The economics literature has also been interested in various logics,
particularly logics of knowledge and belief, and more recently
conditional logic for counterfactual reasoning.  By and large, they
dispense with syntax altogether, working purely at a set-theoretic,
semantic level.  To understand how this is done, it is perhaps best to
take as an example epistemic logic.

Both approaches would start with what is
called a {\em frame\/} in the logic literature, that is, a set $\Omega$
of possible worlds and a binary relation $\K$ on $\Omega$.  Intuitively,
$(w,w') \in \K$ if in world $w$, the agent considers $w'$ possible.%
\footnote{Actually, in the economics literature, it is more standard to
consider a partition of $\Omega$; this is equivalent to the case where
$\K$ is an equivalence relation.}
In the economics literature, the $\K$ relation is used to
define an operator $\Ktt$ on events (subsets of
$\Omega$).  Taking $\K(w) = \{w' : (w,w') \in \K\}$, the operator $\Ktt:
2^\Omega \mapsto 2^\Omega$ is defined as follows:
\begin{equation}\label{introeq}
\Ktt(E) = \{w: \K(w) \subseteq E\}.
\end{equation}
We read $\Ktt(E)$ as ``the agent knows $E$''.

The mathematical/philosophical approach adds an extra level of
indirection to
this more set-theoretic approach.  A language for reasoning
about knowledge is defined, starting in the usual way with a set $\Phi$
of primitive propositions, and closing off under conjunction, negation,
and applications of the modal operator $K$.  As is well known, to give
semantics to formulas in this language, we use a Kripke structure $M =
(W,\K,\pi)$, where $(W,\K)$ is a frame, and $\pi$ is an {\em
interpretation\/} that associates with each primitive proposition and
each world a truth value; that is,
$\pi: \Phi \times \Omega \mapsto \{{\bf true}, {\bf false}\}$.
We then define what it means for a formula $\phi$ to be true at a world
$w \in W$, written $(M,w) \sat \phi$, using the usual inductive
definition.%
\footnote{Readers unfamiliar with the definition can find details in
Section~2.}
We can think of $\sat$ as just associating with each formula $\phi$ an
event $\intension{\phi}_M
= \{w: (M,w) \sat \phi\}$ in structure $M$
($\intension{\phi}_M$ is called the {\em intension\/} of $\phi$ in the
literature \cite{FHMV}).
Not surprisingly,
$\intension{K\phi}_M = \Ktt(\intension{\phi}_M)$;
that is, we obtain the event associated with the formula $K\phi$ by
applying the operator $\Ktt$ to the event associated with $\phi$.

If all that is done with a formula is to translate it to an event, why
bother with the overhead of formulas and $\sat$?  Would it not just be
simpler to dispense with formulas and interpretations, and work directly
with events?  Syntax often plays an important role---for
example, it allows us to express concepts in a model-independent way.  On
the other hand, if we have in mind an intended model all along, then
perhaps it makes sense to just work directly with events.
For example, the {\em model-checking\/} approach, which has been widely
used in proving
correctness of programs \cite{CES}, typically works with one fixed
model, the one generated by the program whose correctness we are trying
to prove.  Model checking  has been advocated for epistemic
reasoning as well \cite{HV91a}.  Perhaps when using the
model-checking approach, it might make sense to work
at the set-theoretic level.

Probability provides another example.  Probabilists start by defining a
particular model---the probability space of interest---and then
investigating its properties.  As the
many texts on probability demonstrate, they have been able to
prove a great many results about probability by working purely at a
set-theoretic level.  While some logics for reasoning about
probability
have been proposed, both propositional \cite{FHM} and first-order
\cite{Hal4}, they certainly do not begin to capture all the subtleties
of the reasoning we find in probability texts.
For example, typical logics of probability cannot express notions
such as expectation and variance.

One of the apparent advantages of working with syntax is that we can
define a proof theory, that allows us to manipulate formulas in order
to prove properties of interest.  We do not have to give up proof theory
if we work at the set-theoretic level.  For example, a standard property
of knowledge is {\em introspection}: if an agent knows a fact, then he
knows that he knows it, and if he does not know it, then he knows that
he does not know it.  Syntactically, these properties are expressed as
\begin{itemize}
\item $K \phi \rimp K K \phi$, and
\item $\neg K \phi \rimp K \neg K \phi$.
\end{itemize}
These properties have immediate set-theoretic analogues:
\begin{itemize}
\item $\Ktt(E) \subseteq \Ktt(\Ktt(E))$, and
\item $\neg \Ktt(E) \subseteq \Ktt(\neg \Ktt(E))$, where $\neg$ here
denotes the set-theoretic complement.
\end{itemize}
As this example suggests, we can translate a syntactic
axiom to a set-theoretic axiom by
\begin{enumerate}
\item replacing formulas by events,
\item replacing the modal operator $K$ by the set operator $\Ktt$,
\item replacing the Boolean operations $\neg, \land, \lor$ by their
set-theoretic analogues $\neg, \inter, \union$.
\end{enumerate}

In this paper, I explore set-theoretic completeness proofs in the
context of epistemic logics and conditional logics.  Both of these
logics were introduced in the philosophical literature
\cite{Hi1,Stalnaker68}, but have been widely used in computer science
and AI.
Epistemic logic has been used as a tool for
analyzing multi-agent systems \cite{FHMV}; conditional logic has been
used as a framework for analyzing nonmonotonic reasoning
\cite{Boutilier94AIJ1} and counterfactual reasoning \cite{Lewis73}.  It
also has
an important role to play in the analysis of causality \cite{Lewis73},
which is becoming an increasingly important issue in AI as well
\cite{Pearl95}.
Set-theoretic completeness proofs for logics of knowledge
and common knowledge are standard in the economics literature (see, for
example, \cite{Aumnotes,Milg}).  I compare them to the
more familiar syntactic completeness proofs in the philosophical
literature, and then do the same for conditional logic.
For the logics considered here, every syntactic operator has a semantic
counterpart; thus, every property expressible syntactically has a
semantic counterpart.  However, as we shall see, the converse is not
always true.  For the logics considered here,
the set-theoretic approach
gives us extra expressive power, allowing us to express more properties.

In part, this comes from the use of arbitrary (rather than
just finite) unions and intersections over events.
We can already see the use of countable intersections and unions in
the context of probability theory.  Although probability is typically
taken to be countably additive, this fact is not expressible in
propositional logics of probability, although it can be expressed once
we allow countable operations.%
\footnote{For those readers unfamiliar with probability, finite
additivity says that if $E$ and $F$ are disjoint sets, then $\Pr(E
\union F) = \Pr(E) + \Pr(F)$.  Countable additivity says that if $E_1,
E_2, \ldots$ is a sequence of pairwise disjoint sets, then $\Pr(\union_i
E_i) = \sum_i \Pr(E_i)$.}
To be precise,
suppose that we have an operator of the form $pr^p(\phi)$ in the
language, which is to be interpreted as ``the probability of $\phi$ is
at least $p$'', and a corresponding operator on events $\mu^p(E)$.
The property of countable additivity cannot be expressed in a
propositional logic, because we do not have countable disjunctions.
Indeed, in \cite{FHM}, a complete axiomatization is given for a logic of
probability where, semantically, probability is taken to be countably
additive, but axiomatically, we require only finite additivity.
(This means that there will be nonstandard models of the theory where
the probability is finitely additive but not countably additive.)  By
way of contrast,
countable additivity is immediately expressible using $\mu^p$ though,
using countable unions:  If $E_i, i \in I$ is a countable
collection of pairwise disjoint sets, then countable additivity just
says
$$\inter_{i\in I} \mu^{p_i}(E_i) \subseteq \mu^p(\union_{i\in I} E_i),
\mbox{where $p = \sum_{i \in I} p_i$.}$$

In the case of knowledge, there is also a property that involves
infinite intersection.  With only finite intersection, we have:
\begin{equation}\label{keq1}
\Ktt(E) \inter \Ktt(E') = \Ktt(E \inter E').
\end{equation}
Once we allow infinite intersections, we have
\begin{equation}\label{keq2}
\mbox{$\inter_{j \in J}\Ktt_i(E_j) = \Ktt_i(\inter_{j \in J} E_j)$
for any index set $J$ and events $E_j$, $j \in J$.}
\end{equation}
Clearly (\ref{keq2}) implies (\ref{keq1}).  Somewhat surprisingly, it
can be shown that if $\K$ is an equivalence relation (that is, knowledge
satisfies the properties of {\bf S5}), then they are equivalent.  (This
follows from Aumann's set-theoretic completeness proof \citeyear{Aumnotes},
although I provide a direct proof.)
However, once we weaken the
{\bf S5} properties, for example, if we consider either {\bf S4} or {\bf
K45}, then the
equivalence no longer holds, and
we need the full strength of (\ref{keq2})
to get set-theoretic completeness proofs.  At the syntactic level, this
distinction is lost, because infinite conjunctions cannot be expressed.

Of course, it can be argued that if we extended propositional logic to
allow infinite conjunctions, then these differences could be expressed
perfectly well syntactically.
However, issues of expressiveness do not arise only for infinite
conjunctions and
disjunctions.  In the context of conditional logic, I show that there
are properties that involve only finite intersections and unions that
have
no analogue at the syntactic level.  Moreover, even properties that can
be captured syntactically are more naturally expressed at the
set-theoretic level.  Finally, as we shall see, completeness proofs seem
to be more straightforward and transparent at the set-theoretic level,
at least in the case of the logics considered here.

The rest of this paper is organized as follows.  In
Section~\ref{epistemic}, I consider epistemic logic, while in
Section~\ref{conditional}, I consider conditional logic.
I conclude in Section~\ref{conclusion}.

\section{Epistemic Logic}\label{epistemic}
I start by reviewing the syntactic approach to epistemic logic, and then
I examine how the set-theoretic approach works.

\subsection{The Syntactic Approach: A Review}
For simplicity here, I consider single-agent epistemic logic; all the
points I want to make already arise in the single-agent case.  I briefly
review the syntax and semantics here for those not familiar with it.
We start with a nonempty set $\Phi$ of primitive propositions, and close
off under negation, conjunction, and applications of the modal operator
$K$.  Let $\LPK$ be the language consisting of all formulas that
can be built up this way.  Thus, a typical formula in $\LPK$ is
$K (\neg K(p\land q))$. We define implication and disjunction as usual.

A {\em Kripke structure\/} is a tuple $M = (W,\K,\pi)$, as discussed in
the introduction.  We define $(M,w) \sat \phi$ by induction as follows:
\begin{description}
\item
$(M,\world)\sat p$ (for a primitive proposition
$p\in\Phi$) iff  $\pi(\world,p)={\bf true}$
\item $(M,\world)\sat\phi\land \phi'$  iff  $(M,\world)\sat \phi$ and
$(M,\world)\sat \phi'$
\item
$(M,\world)\sat \neg \phi$ iff $(M,\world)\not\sat \phi$
\item $(M,\world)\sat K\phi$ iff  $(M,\world')\sat \phi$ for all
$\world'$ such that $(w,w') \in \K$.
\end{description}

There are well-known soundness and completeness results for
epistemic logic, that show the close connection between the assumptions
we make about $\K$ and axiomatic properties.  Consider the following
axioms:
\begin{itemize}
\item[Prop.]
All substitution instances of tautologies of propositional calculus
\item[K1.]
$(K\varphi\land K(\varphi\rimp  \psi)) \rimp
K\psi$,\   (Distribution Axiom)
\item[K2.] $K \phi \rimp \phi$,  \  (Knowledge
Axiom)
\item[K3.] $K \phi \rimp K K \phi$,  \
(Positive Introspection Axiom)
\item[K4.] $\neg K \phi \rimp K \neg K \phi$,
\ (Negative Introspection Axiom)
\item[MP.]
From $\varphi$ and $\varphi\rimp\psi$
infer $\psi$\ \  (Modus ponens)
\item[Gen.]
From $\varphi$ infer $K \varphi$\ \
(Knowledge Generalization)
\end{itemize}

The system with axioms and rules Prop, K1, MP, and Gen has been
called~{\bf K}.  If we add K2 to {\bf K}, we get {\bf T}; if we add K3
to {\bf T}, we
get {\bf S4}; if we add K4 to {\bf S4}, we get {\bf S5}; finally, if we
add K3 and K4 to
{\bf K}, we get {\bf K45}.  (Other systems can also be formed; these are
the ones I focus on here.)

Let $\M$  be the class of all Kripke structures.  We are also interested
in subclasses of $\M$ where the $\K$ relation has various properties of
interest.  Let $\M^r$
(\respc $\M^{rt}$; $\M^{et}$; $\M^{rst}$) consist of the Kripke
structures where the $\K$ relation is reflexive (\respc reflexive and
transitive; Euclidean%
\footnote{A relation $R$ is Euclidean if $(s,t), (s,u) \in R$ implies
that $(t,u) \in R$.}
and transitive; reflexive, symmetric, and transitive, \ie an equivalence
relation).

\thmm\label{thm5} For formulas in the language $\LPK$:
\begin{enumerate}
\item[(a)] {\bf K} is a sound and complete axiomatization with respect
to $\M$,
\item[(c)] {\bf T}  is a sound and complete axiomatization with
respect to $\M^r$,
\item[(c)] {\bf S4}  is a sound and complete axiomatization with
respect to $\M^{rt}$,
\item[(d)]  {\bf K45} is a sound and complete axiomatization with
respect to $\M^{et}$.
\item[(e)] {\bf S5} is a sound and complete axiomatization with
respect to $\M^{rst}$.
\end{enumerate}
\ethm

\subsection{The Set-Theoretic Approach}
In the set-theoretic approach, we just start with a frame $(W,\K)$.  We
can then define an operator $\Ktt$ as in Equation~\ref{introeq} in the
Introduction.  Consider the
following properties of the $\Ktt$ operator:

\begin{itemize}
\item[A1.] $\Ktt(E) \inter \Ktt(E') = \Ktt(E \inter E')$
\item[A2.] $\Ktt(E) \subseteq E$
\item[A3.] $\Ktt(E) \subseteq \Ktt(\Ktt(E))$
\item[A4.] $\neg \Ktt(E) \subseteq \Ktt(\neg \Ktt(E))$
\item[A5.]
$\inter_{j \in J} \Ktt(E_j) = \Ktt(\inter_{j \in J} E_j)$
for any index set $J$ and events $E_j, j \in J$%
\footnote{If $J = \emptyset$, we take the intersection over the empty
set to be $\Omega$, as usual.  Thus, as a special case of this axiom, we
get $\Omega = \Ktt(\Omega)$.}
\end{itemize}
A2, A3, and A4 are the obvious analogues of K2, K3, and K4,
respectively.  A1 can be viewed as an analogue of K1.  In fact, it is
not hard to show that
\begin{itemize}
\item[K1$'$.] $K(\phi \land \psi) \dimp (K\phi \land K\psi)$
\end{itemize}
is equivalent to K1 in the presence of MP and Prop.  We
could have replaced K1 by K1$'$ in all the axiom systems and still have
obtained all the completeness proofs of Theorem~\ref{thm5}.
Instead of A1, Aumann uses the monotonicity property
\begin{itemize}
\item[A1$'$.] If $E \subseteq F$, the $\Ktt(E) \subseteq \Ktt(F)$.
\end{itemize}
It is easy to see that A1$'$ follows from A1 (since if $E \subseteq F$,
then $E \inter F = E$, so $\Ktt(E) \subseteq \Ktt(E) \inter \Ktt(F)
\subseteq \Ktt(F)$).  A1 does not follow from A1$'$, but it follows from
Aumann's set-theoretic completeness theorem that A1 does
follow from A1$'$, A2, A3, and A4; Proposition~\ref{A1'} provides
a self-contained proof of this fact.

Note there
is no analogue to Prop, MP, or Gen above; they turn out to be
unnecessary at the set-theoretic level.  (In particular, once we work at
the level of sets, we do not need the Boolean equivalences encoded in
Prop as axioms.)  On the other hand, there is no analogue of A5 at
the syntactic level; we cannot express infinite conjunctions in
propositional logic, so it is unnecessary.
It turns out that A5 is also unnecessary if we assume that $\K$ is an
equivalence relation.  This follows from the following result.

\prom\label{A1'} Any operator on events in $\Omega$ that satisfies
A1$'$, A2, and A4 must also satisfy A5 (and hence A1). \epro
\prf Suppose $\Ktt$ satisfies A1$'$, A2, and A4.  Consider the fixed
points of
$\Ktt$, that is, those sets $E$ such that $\Ktt(E) = E$.  I first show
that the set of fixed points of $\Ktt$ is closed under negations and
arbitrary unions.

Suppose $\Ktt(E) = E$.  Then $\neg E = \neg \Ktt(E)$, so
$\Ktt(\neg E) = \Ktt(\neg
\Ktt(E)) = \neg \Ktt(E) = \neg E$, where the second equality follows
from A2 and A4.  Thus, $\neg E$ is a fixed point of $\Ktt$.

Next, suppose $\Ktt(E_j) = E_j$ for all $j$ in some index set
$J$.  Since $E_j \subseteq \union_j E_j$, by A1$'$,
we have $\Ktt(E_j) \subseteq K(\union_j E_j)$.
Thus,
\begin{equation}\label{keq3}
\union_j \Ktt(E_j) \subseteq \Ktt(\union_j E_j).
\end{equation}
It now follows that
$$\begin{array}{lll}
&\Ktt(\union_{j\in J} E_j)\\
= &\Ktt(\union_{j \in J} \Ktt(E_j)) &\mbox{since $E_j = \Ktt(E_j)$}\\
\subseteq &\Ktt(\Ktt (\union_{j \in J} E_j)) &\mbox{by A1$'$ and
(\ref{keq3})}\\
\subseteq &\Ktt(\union_{j \in J} E_j) &\mbox{by A2.}
\end{array}$$
Thus, $\Ktt(\union_{j \in J} E_j) = \Ktt\Ktt(\union_{j \in J} E_j)$, and
the set of fixed points of $\Ktt$ is closed under arbitrary
unions.

We are now ready to prove A5.  Since $\inter_{j \in J} E_j \subseteq
E_j$, it follows from A1$'$ that
$\Ktt(\inter_{j \in J} E_j) \subseteq \Ktt(E_j)$.  Thus,
$\Ktt(\inter_{j \in J} E_j) \subseteq \inter_{j \in J} \Ktt(E_j)$.  For
the opposite inclusion, observe that by A2, we have
$\inter_{j \in J} \Ktt(E_j) \subseteq \inter_{j \in J} E_j$.  Thus, by
A1$'$,
$\Ktt(\inter_{j \in J} \Ktt(E_j)) \subseteq \Ktt(\inter_{j \in J} E_j)$.
By A2 and A4, $\neg \Ktt(E_j)$ is a
fixed point of $\Ktt$, for all $j \in J$.  Since the set of fixed points
is closed under
arbitrary unions and negations, $\inter_{j\in J} \Ktt(E_j)$ is also a
fixed point of $\Ktt$. It follows that
$\inter_{j \in J} \Ktt(E_j) \subseteq \Ktt(\inter_{j \in J} E_j)$, as
desired. \eprf

Each of A1$'$, A2, and A4 is necessary for Proposition~\ref{A1'}.  If
we drop any of them, then A5 no longer necessarily holds, as the
following examples show.

\xam Let $\Omega = \{1,2,3\}$ and define $\Ktt_0(\{1\}) =
\Ktt_0(\{2,3\}) = \emptyset$, and $\Ktt_0(E) = E$ for $E \ne \{1\},
\{2,3\}$.  It is easy to see that $\Ktt_0$ satisfies A2 and A4, but not
A1$'$ (since $\{3\} \subseteq \{2,3\}$ but $\Ktt_0(\{3\}) = \{3\}
\not\subseteq \emptyset = \Ktt_0(\{2,3\})$).  Since $\Ktt_0$ does not
satisfy A1$'$, {\em a fortiori}, it does not satisfy A1 or A5.
\exam

\xam  Let $\Omega = \{1,2,3, \dots\}$.
Define $\Ktt_1(E) = E$ if $E$ is
cofinite (that is, if the complement of $E$ is finite) and $\Ktt_1(E) =
\emptyset$ otherwise.  It is easy to see that $\Ktt_1$ satisfies A1
(and hence A1$'$) and
A2, but does not satisfy A4 (since, for example, $\neg \Ktt_1( \neg
\{1\}) = \{1\} \ne \emptyset = \Ktt(\neg \Ktt (\neg \{1\})$).  $\Ktt_1$
does not satisfy A5, since if $E_j = \neg \{j\}$, then $\Ktt_1(\inter_{j
\ge 1} E_j) = \Ktt_1(\{1\}) = \emptyset \ne \{1\} = \inter_{j \ge 1}
\Ktt_1(E_j)$. \exam

\xam Let $\Omega = \{1,2,3, \ldots\}$.
Define $\Ktt_2(E) = \Omega$ if $E$ is cofinite and
$\Ktt_2(E) = \emptyset$ otherwise.  Again, it is easy to see that
$\Ktt_2$ satisfies A1 and A4, but does not satisfy A2 (since, for
example, $\Ktt_2(\neg\{1\}) = \Omega \not\subseteq \neg\{1\}$).  Taking
$E_j = \neg\{j\}$ as in the previous example, note that
$\Ktt_2$ does not satisfy A5 since $\Ktt_2(\inter_j E_j) = \emptyset
 \ne \Omega = \inter_j \Ktt_2(E_j)$. \exam

The following theorem is the set-theoretic analogue of
Theorem~\ref{thm5}.  Aumann \citeyear{Aumnotes} proved it for the case
that $M \in \M^{rst}$ and $\Ktt$ satisfies A1$'$, A2--A4; this is a
generalization of his result.  (In light of Proposition~\ref{A1'}, we
can replace A1$'$ by A1.)  It is just the result we would expect
(modulo, perhaps, the need for A5 if we do not have all of A1, A2, and
A4).

\thmm\label{completeness}  The $\Ktt$ operator in the frame $(\Omega,\K)$
satisfies A5.  Moreover, if $\K$ is reflexive (\respc reflexive and
transitive; Euclidean and transitive; an equivalence relation) then
$\Ktt$ satisfies A2 (\respc A2 and A3; A3 and A4; A1--A4).  Conversely,
if $\Ktt'$ is an operator on events satisfying A5, then there is a
binary relation $\K$ on $\Omega$ such that $\Ktt'$ is the $\Ktt$ relation in
the frame $(W,\K)$.  Moreover, if $\Ktt'$ satisfies
A2 (\respc A2 and A3; A3 and A4; A1--A4), then $\K$ is
reflexive (\respc reflexive and transitive; Euclidean and transitive; an
equivalence relation).
\ethm

\prf The first part is straightforward and left to the reader.  For the
second part, given an operator $\Ktt'$ on $\Omega$, define
$\K$ so that $\K(w) = \inter \{E: w\in \Ktt'(E)\}$.  Using
the fact that in all cases $\Ktt'$ satisfies A5, it is easy to check
that $\Ktt$ and $\Ktt'$ agree.  And just as in the standard canonical
model constructions of completeness in the syntactic case (see, \eg
\cite{FHMV}), we can show that A2 forces $\K$ to be reflexive, A3 forces
it to be transitive, and A4 forces it to be Euclidean.  The result
follows.  (Note that a reflexive, Euclidean, and transitive relation is
an equivalence relation.)
\eprf

This theorem is a
typical example of a set-theoretic soundness and completeness result.
The first part can be viewed as a soundness statement, while the second
part gives us completeness.

The reader familiar with the standard syntactic completeness proofs using
canonical model constructions should find it instructive to compare the
set-theoretic completeness proofs with those involving canonical models.
The set-theoretic proofs works for an arbitrary set $\Omega$ of worlds;
we do not have to construct a special set where each world corresponds
to a maximal consistent set of formulas.  The definition of the $\K$
relation above is very similar in spirit to that in the canonical model
construction, as are the arguments that A2, A3, and A4 force the $\K$
relation to be reflexive, transitive, and Euclidean, respectively.
However, the proof that $\Ktt = \Ktt'$ is simpler than the proof
that a formula is true at a world in the canonical model if and only if
it is an element of that world (viewed as a maximal consistent set of
formulas).  As we shall see, in the case of conditional logic,
set-theoretic completeness proofs are also relatively simpler than
syntactic ones.

I have said we can view Theorem~\ref{completeness} as a set-theoretic
soundness and completeness result.  The standard
soundness and completeness results in logic involve a language, a proof
theory, and a semantics.  Is there a way we can view
Theorem~\ref{completeness} as a more standard soundness and completeness
result?  I briefly sketch here an argument showing that we can.

Fix a finite set of worlds $W_0$.
The set of {\em event
descriptions (for $W_0$)\/} is the least set formed as follows: We have
a symbol
${\sf A}$ for each subset $A$ of $W_0$, and
close off under union, complementation, and
applications of the $\Ktt$ operator.  Of course, we take ${\sf
E}_1 \inter {\sf E}_2$ to be an abbreviation for $\neg( \neg
{\sf E}_1 \union \neg {\sf E}_2)$. A
{\em basic event formula (for $W_0$)\/} has the form
${\sf E}_1 = {\sf E}_2$, where ${\sf E}_1$ and
${\sf E}_2$ are event descriptions.  Note that ${\sf E}_1 \subseteq
{\sf E}_2$ can be viewed as an abbreviation for the basic event
formula ${\sf E}_1 \inter {\sf E}_2 = {\sf E}_1$.  An {\em event
formula (for $W_0$)\/} is a Boolean combination of basic event formulas.
Our language consists of event formulas.  Note that the language is
relative to the particular domain $W_0$ about which we wish to reason.

A semantic model for this language consists of an interpretation of
$\Ktt$ as an operator on subsets of $W_0$.  This allows us to associate
with each event description ${\sf E}$ a subset $v({\sf E})$ of $W_0$ in
the obvious way.  Of course, each symbol
${\sf A}$ is interpreted as the corresponding subset $A$ of $W_0$ (that
is, $v({\sf A}) = A$).  Union and complementation get there standard
interpretation, and the interpretation of $\Ktt({\sf E})$ is determined
by the
interpretation of $\Ktt$.  Finally, a basic event formula ${\sf E}_1 =
{\sf E}_2$ is true relative to an interpretation if $v({\sf
E}_1) = v({\sf E}_2)$.  Boolean combinations of event formulas are
interpreted in the obvious way.

As for the axiom system, besides considering A1--A4 (or some subset of
these axioms), we need Prop and MP from the axiom system {\bf K}
for propositional reasoning,
an axiom that says $\Ktt({\sf E})$ is equal to
some subset of $W_0$, axioms describing the relationship between subsets
of $W_0$, and axioms and inference rules for dealing with equality.
The axiom that says $\Ktt({\sf E})$ is equal to some subset of $W_0$ is
simply
\begin{itemize}
\item[A6.] $\union_{A \subseteq W_0} \Ktt({\sf E}) = {\sf A}.$
\end{itemize}
The axioms describing the relationship between subsets of $W_0$ have the
following form: for all subsets $A, B, C \subseteq W_0$, if $A \union B =
C$,
then we have an axiom ${\sf A} \union {\sf B} = {\sf C}$; similarly, if
$B = \neg A$, we have the axiom ${\sf B} = \neg {\sf A}$.  Call this
collection of axioms Rel.

Finally, we need axioms that say equality is an equivalence relation
(reflexive, symmetric, and transitive), and an inference rule that
allows us to substitute equals for equals.  Call these three axioms and
inference rule Eq.

Let ${\cal A}$ be the axiom system consisting of A1, A6, Prop, MP, Rel,
and Eq.

\corm\label{completecor} Let ${\cal A'}$ be a subset of A2, A3, A4.  Then
${\cal A}
\union
{\cal A'}$ is a sound and complete axiomatization for the set of frames
of the form $(W_0,\K)$ where $\K$ satisfies the subset of $\{$reflexive,
transitive, Euclidean$\}$ corresponding to ${\cal A'}$. \ecor

\prf Soundness is obvious.  For completeness, it suffices to show that
if $\phi$ is an event formula that is consistent with ${\cal A} \union
{\cal A'}$, then $\phi$ is satisfied in a frame $(W_0,\K)$ where $\K$
satisfies the appropriate properties.  But this follows almost
immediately from Theorem~\ref{completeness}.  Extend $\phi$ to a maximal
complete subset of event formulas.  This set of formulas defines an
operator $\Ktt$ on events satisfying A1 (and hence A5, since $W_0$ is
finite). Thus, by Theorem~\ref{completeness}, there is a binary relation
$\K$ on $W_0$ with the right properties.  \eprf

If $W_0$ is infinite rather than finite, it seems that to prove a result
like Corollary~\ref{completecor}, we need to allow arbitrary unions
rather than just finite unions and arbitrary disjunctions (to express
the analogue of A6).  I conjecture that a completeness proof exists even
if we restrict the language to finite disjunctions, although I have not
checked details.  In any case, Corollary~\ref{completecor} does show
that it is legitimate to view Theorem~\ref{completeness} as a semantic
counterpart of the more usual soundness and completeness results.%
\footnote{I thank Giacomo Bonanno for raising this issue.}

\section{Conditional logic}\label{conditional}

As I said earlier, conditional logic has been used to capture both
counterfactual reasoning and default reasoning.
Stalnaker \citeyear{Stalnaker92a} gives a
short and readable survey of the philosophical issues involved.  I
briefly review the standard syntax and semantics here.

\subsection{The Syntactic Approach: Selection Functions}
As suggested above, the syntax for conditional logic is straightforward.
We start with a set $\Phi$ of primitive propositions, and close it off
under conjunction, negation, and applications of the binary modal
operator $\Cond$.  Thus, if $\phi$ and $\psi$ are formulas, then so is
$\phi \Cond \psi$.  The formula
$\phi \Cond \psi$, can be read as ``if $\phi$ were
the case, then $\psi$ would be true'' (if we want to give $\Cond$ a
counterfactual reading) or ``typically/normally/by default, if $\phi$ is
the case then $\psi$ is the case'' (if we want to give it a reading more
appropriate for nonmonotonic reasoning).
Let $\LPC$ be the set of formulas that can be built up this way.

The original approach for capturing $\Cond$, due to Stalnaker
and Thomason \cite{Stalnaker68,StalThom70}, proceeds as follows:
Given a language $\LPC$, they assume that there is
a {\em selection function\/} $f: \Omega \times \LPC
\mapsto \Omega$.  For counterfactual reasoning, we can think of
$f(\omega,\phi)$ as the world ``closest'' to $\omega$ that satisfies
$\phi$.
For default reasoning, $f(\omega,\phi)$ should be thought of as the most
normal world (relative to $\omega$) satisfying $\phi$.  These
interpretations implicitly
assume that there is a unique world closest to $\omega$ (or most
normal relative to $\omega$) that satisfies
$E$.  Many later authors argued that there is not in general a
unique closest world; ties should be allowed.
I follow this interpretation here.
Thus, I take a {\em
counterfactual structure\/} to be a tuple $(\Omega,f,\pi)$, where $f:
\Omega \times \LPC \mapsto 2^\Omega$
and $\pi$ is an interpretation, as before.

The definition of $\sat$ in counterfactual structures is
the same as that in epistemic structures,
except for the clause for $\Cond$, which is
\begin{description}
\item $(M,w) \sat \phi \Cond \psi$ iff
$f(w,\phi) \subseteq \intension{\psi}_M$.
\end{description}
This captures the intuition that the closest worlds to $w$, as defined
by $f$, satisfy $\phi$.

There are various restrictions we can consider placing on the selection
function.
\begin{itemize}
\item[S1.] $f(\omega,\phi) \subseteq \intension{\phi}_M$: the worlds
closest to $\omega$ satisfying $\phi$ are in fact $\phi$-worlds
(where $w$ is a {\em $\phi$-world\/} if $w \in \intension{\phi}_M$).
\item[S2.] If $f(\omega,\phi) \subseteq \intension{\psi}_M$ and
$f(\omega,\psi) \subseteq \intension{\phi}_M$, then $f(\omega,\phi) =
f(\omega,\psi)$.  Stalnaker and Thomason \citeyear{StalThom70} view this
as a {\em uniformity\/} condition.  If the closest $\phi$-worlds all
satisfy $\psi$ and the closest $\psi$-worlds all satisfy $\phi$, then
the closest $\phi$-worlds and the closest $\psi$-worlds must be the
same.  Note that S1 and S2 together force $f$ to be a semantic function:
if $\intension{\phi}_M = \intension{\psi}_M$, then, in the presence of
S1, the antecedent to S2 will hold, so $f(w,\phi) = f(w,\psi)$.
\item[S3.] If $\omega \in \intension{\phi}_M$, then $f(\omega,\phi)=
\{\omega\}$: if
$\omega$ is a $\phi$-world, then it is the closest $\phi$-world to
$\omega$.
This restriction is particularly appropriate for counterfactual
reasoning.  It is not necessarily appropriate for nonmonotonic
reasoning.  The most normal $\phi$-world may not be
$w$, even if $w$ is a $\phi$-world.
\item[S4.] $f(\omega,\phi)$ is either empty or a singleton.
This captures Stalnaker's original assumption that there is a unique
closest world, if there is a closest world at all.
\item[S5.] $f(\omega,\phi_1 \lor \phi_2) \subseteq f(\omega,\phi_1)
\union
f(\omega,\phi_2)$: if $\omega'$ is one of the $(\phi_1 \lor
\phi_2)$-worlds closest to $\omega$, then it must be one of the
$\phi_1$-worlds closest to $\omega$ or one of the $\phi_2$-worlds
closest to $\omega$.
\item[S6.] If $f(\omega, \phi) \subseteq \intension{\psi}_M$ then
$f(\omega, \phi \land \psi)
\subseteq f(\omega,\phi)$: if the closest
$\phi$-worlds to $\omega$ all satisfy $\psi$, then the closest $\phi
\land \psi$-worlds are all among the closest $\phi$-worlds to $\omega$.%
\footnote{It may seem even more reasonable to replace $\subseteq$ by
$=$ here.  This would say that if the closest $\phi$-worlds to $\omega$
satisfy $\psi$, then they are the closest $\phi \land \psi$-worlds.
In fact, the stronger version of S6 already follows
from S1, S2, and S5.  (Proof: By S1, S2, and S5, $f(\omega,\phi)
= f(\omega,(\phi \land \psi) \lor (\phi \land \neg\psi)) \subseteq
f(\omega,\phi \land \psi) \union f(\omega, \phi \land \neg \psi)$.
Moreover, if $f(\omega,\phi)
\subseteq \intension{\psi}$, it follows from S1 that $f(\omega,\phi)
\inter f(\omega,\phi \land \neg \psi) = \emptyset$.  Thus,
$f(\omega,\phi) \subseteq f(\phi \land \psi)$.)  Note that S2 can
easily be obtained from S1 and the stronger version of S6.
I use the weaker version of S6 here because it allows
us to make technical connections to some known results in conditional
logic.}
\item[S7.] If $f(\omega,\phi) \inter \intension{\psi}_M \ne \emptyset$,
then $f(\omega, \phi \land \psi)
\subseteq f(\omega,\phi) \inter \intension{\psi}_M$: this is a
strengthening of S5 (at least, when $f(\omega,\phi) \ne \emptyset$),
which
says that the closest $\phi\land \psi$-worlds to $\omega$ are among the
closest $\phi$-worlds that are also in $\intension{\psi}_M$ (provided
there are any).%
\footnote{Again, we may want to strengthen $\subseteq$ to $=$, and
again, the stronger version follows from the weaker version, in the
presence of S1, S2, and S5.}
\item[S8.] If $\intension{\phi}_M \ne \emptyset$ then $f(\omega,\phi)
\ne \emptyset$: this
says that there always is some $\phi$-world closest to $w$ if there are
any $\phi$-worlds.
\end{itemize}
As we shall see in Section~\ref{ordering}, if we introduce an
ordering on worlds and define $f(w,\phi)$ as the $\phi$-worlds closest
to $w$, then these restrictions arise by taking some very natural
restrictions on the ordering (indeed, restrictions S1, S2, S5, and S6
are forced on us).

There is a well-known axiom corresponding to each of these conditions
except for S8. Let $\Box \phi$ be an abbreviation for $\neg \phi \Cond
\false$
and let $\Diamond \phi$ be an abbreviation of $\neg \Box \neg \phi$.
Thus, $(M,w) \sat \Box \phi$ iff $f(w,\neg \phi) = \emptyset$ and
$(M,w) \sat \Diamond \phi$ iff $f(w,\phi) \ne \emptyset$.
Consider the following axioms:

\begin{itemize}\denselist
 \item[Prop.] All substitution instances of propositional tautologies
 \item[C0.] $((\phi \Cond \psi_1)\land(\phi\Cond\psi_2)) \rimp
    (\phi\Cond(\psi_1\land\psi_2))$
 \item[C1.] $\phi \Cond \phi$
\item[C2.]  $((\phi \Cond \psi) \land (\psi \Cond \phi)) \rimp ((\phi
\Cond \sigma) \rimp (\psi \Cond \sigma))$
 \item[C3.] $\phi \rimp (\psi \dimp (\phi \Cond \psi))$
 \item[C4.] $(\phi \Cond \psi) \lor (\phi \Cond \neg \psi)$
 \item[C5.] $((\phi_1\Cond\psi)\land(\phi_2\Cond\psi)) \rimp
    ((\phi_1\lor\phi_2) \Cond\psi)$
 \item[C6.] $((\phi_1\Cond\phi_2)\land(\phi_1\Cond\psi)) \rimp
    ((\phi_1\land\phi_2) \Cond \psi)$
 \item[C7.] $(\neg (\phi_1\Cond \neg \phi_2)\land(\phi_1\Cond\psi))
\rimp ((\phi_1\land\phi_2) \Cond \psi)$
\item[C8.]
(a) $\Box \phi \rimp (\phi \land (\phi' \Cond \Box \phi))$\\
(b) $\Diamond \phi \rimp (\phi' \Cond \Diamond \phi)$
 \item[MP.] From $\phi$ and $\phi \rimp \psi$ infer $\psi$
 \item[RC1.] From $\psi \rimp \psi'$ infer $(\phi\Cond\psi) \rimp
(\phi\Cond\psi')$
\end{itemize}

All of the axioms
other than C8
above
are familiar from the literature.
In the language of \cite{KLM}, C0 is the And Rule, C1 is Reflexivity, C5
is the Or Rule, C6 is Cautious Mononotonicity, C7 is Rational
Monotonicity, and RC1 is Right Weakening.
In \cite{KLM} the focus was default reasoning, for which C0, C1, C5, C6,
C7, and RC1 are considered appropriate.
C2,
C3, and C4 come from the
literature on counterfactual reasoning; they are
A6, t4.9, and
t4.7 in \cite{StalThom70}, respectively.
There is one other rule considered in \cite{KLM} called Left Logical
Equivalence, which says
\begin{itemize}
\item[LLE.] From $\phi \dimp \phi'$ infer $(\phi\Cond\psi) \rimp
(\phi'\Cond\psi)$
\end{itemize}
This is true if $f$ is a semantic notion, which depends only on
$\intension{\phi}_M$, and not the syntactic form of $\phi$.  As we have
observed, this follows from S1 and S2.  Not surprisingly, LLE follows
readily from C1, C2, and RC1.  We can also obtain C2 from C0, C1, C5,
C6, RC1, and LLE, with a little effort, thus C2 holds in the framework of
\cite{KLM}.

C8 is intended to characterize S8.  Basically,
we want to say that if $\intension{\phi}_M \ne \emptyset$ (\ie $\phi$ is
satisfiable somewhere in structure $M$) then $\Diamond  \phi$
must be valid in $M$ (that is, true at every world in $M$).
Equivalently,
if $M \sat \Box \phi$, then $\neg \phi$ must be unsatisfiable in $M$.
The following discussion may help explain how C8 captures this.

Given a structure $M = (\Omega,f,\pi)$, define
$w'$ to be {\em reachable from $w$ via $\phi$\/} if $w' \in
f(w,\phi)$.   We can then inductively define reachability via a sequence
$\phi_1; \ldots; \phi_k$.  If $k > 1$, $w'$
{\em is reachable from
$w$ via $\phi_1; \ldots; \phi_k$\/}
if there exists $w''$ such that $w'$ is
reachable from $w''$ via $\phi_k$ and $w''$ is reachable from $w$ via $\phi_1;
\ldots; \phi_{k-1}$.
Finally, we say that $w'$ is {\em reachable from $w$\/} if
it is reachable via some sequence of formulas.  Notice that
$(M,w) \sat \phi_1 \Cond (\phi_2 \Cond ( \ldots (\phi_k \Cond
\phi)\ldots ))$ if
$\phi$ is true at every world $w'$ reachable from $w$ via
$\phi_1; \ldots; \phi_k$.

\lemm\label{C8}  Suppose C8 is valid in $M$.
\begin{itemize}
\item[(a)] If $(M,w) \sat \Box \phi$, then $(M,w') \sat \phi \land \Box
\phi$ for every $w'$ reachable from $w$ via any sequence of formulas.
\item[(b)] If $(M,w) \sat \Diamond \phi$,
then $(M,w') \sat \Diamond \phi$
for every $w'$ reachable from $w$ via any sequence of formulas.
\end{itemize}
\elem
\prf  For part (a), note that
part (a) of C8 says that if $(M,w) \sat \Box \phi$, then
$(M,w') \sat \Box \phi$
for every world $w'$ reachable from $w$ via
$\phi'$.  Inductively, it follows that $(M,w') \sat \Box \phi$
for every world
$\omega'$ reachable from $w$ via any sequence of
formulas.  Since C8 also tells us that $\Box \phi \rimp
\phi$ is valid, this means that $\phi$ is true at all worlds
reachable from $w$ via any sequence of formulas.
Part (b) follows similarly from part (b) of C8.  \eprf

As the following theorem shows, the observation in Lemma~\ref{C8}
is enough to essentially force S8.%
\footnote{
In the presence of C1, C5, C6, RC1, and LLE, C8 can be expressed in more
familiar terms.
It is easy to see that in the presence of C1, C5, C6, RC1, and LLE,
$\Box \phi \rimp (\psi \Cond \phi)$ is valid.  (Proof: From C1, we have
$(\phi \land \psi) \Cond (\phi \land \psi)$.  By RC1, we have
$(\phi \land \psi) \Cond \phi$.  From $\neg \phi \Cond \false$ ($\Box
\phi$) and RC1 we have $\neg \phi \Cond \phi$ and $\neg \phi \Cond\psi$.
From C6, we get that $\neg \phi \land \psi \Cond \phi$.  Now by C5, we
get $((\neg \phi \land \psi) \lor (\phi \land \psi)) \Cond\phi$.
Finally, by LLE, we get $\psi \Cond \phi$.)  Thus,
C8 follows from $(\Box \phi \rimp (\Box \Box \phi \land \phi)) \land
(\Diamond \phi \rimp \Box \Diamond \phi)$.  This is a conjunction of two
axioms denoted R (for reflexivity) and U (for uniformity) by Lewis
\citeyear{Lewis73}.  The converse holds as well; this follows from
Theorem~\ref{condcomplete}.}

\thmm\label{condcomplete} Let $\S$ be a
(possibly empty) subset of $\{$S1, \ldots, S8$\}$ and let
$\C$ be the corresponding subset of $\{$C1, \ldots, C8$\}$.  Then
$\{$\mbox{Prop,C0,MP,RC1}$\} \union \C$ is a sound and complete
axiomatization for the language $\LPC$ with respect to the class of
counterfactual structures satisfying the conditions in $\S$.  \ethm

\prf As usual, soundness is straightforward.  While the basic ideas of
the completeness proof are standard (and go back to
\cite{StalThom70}), I sketch some of the details here, since this result
is somewhat more general than the ones that appear in the literature.
I assume that the reader is familiar
with the standard canonical model arguments from the literature.

For completeness, we must show that every consistent formula is
satisfiable.  Suppose that $\phi_0$ is consistent.  We first consider
the case that $\mbox{C8} \notin \C$.

Let $\Omega$ consist of all the maximal consistent sets of formulas in
$\LPC$.
Let $g: \Omega \times \LPC \mapsto 2^{\LPC}$ be
defined via $g(\omega,\phi) = \{\psi: \phi \Cond \psi \in \omega\}$.
Define a selection $f$ on $\Omega$ so that
$f(\omega,\phi) = \{w' : g(\omega,\phi) \subseteq w'\}$, and
define an interpretation $\pi$ such that $\pi(w,p) = {\bf true}$ iff $p
\in w$.  Let $M = (\Omega,f,\pi)$.

We can now prove the usual {\em Truth Lemma}: for every formula $\phi
\in
\LPC$, we have $(M,w) \sat \phi$ iff $\phi \in w$.  We
proceed by induction; the
only nontrivial case comes if $\phi$ has the form $\psi_1 \Cond \psi_2$.
If $\psi_1 \Cond \psi_2 \in w$, we want to show that $(M,w) \sat
\psi_1\Cond\psi_2$.  Thus, we must show that
$f(w,\psi_1) \subseteq \intension{\psi_2}_M$.
  By definition, $f(w,\psi_1) = \{w': g(w, \psi_1)
\subseteq w'\}$.  Since $\psi_1 \Cond \psi_2 \in w$, it follows that
$\psi_2 \in g(w,\psi_1)$, so $f(w,\psi_1) \subseteq \{w': \psi_2 \in
w'\}$.  By the induction hypothesis, $\{w': \psi_2 \in
w'\} = \intension{\psi_2}_M$.  Thus, $(M,w) \sat \psi_1 \Cond \psi_2$,
as desired.

For the opposite direction, suppose $(M,w) \sat \psi_1 \Cond \psi_2$.
Thus, $f(w,\psi_1) \subseteq \intension{\psi_2}_M$.
It follows that $\psi_2$ must be provable from the formulas in
$g(w,\psi)$,
for otherwise there would be a world $w'$ containing $g(w,\psi_1)$ and
$\neg\psi_2$, and $w'$ would be in
$f(w,\intension{\psi_1}_M) - \intension{\psi_2}_M$, by the induction
hypothesis. Thus, there is a finite set
of formulas in $g(w,\psi_1)$, say $\{\sigma_1, \ldots,\sigma_k\}$, such
that $\sigma_1\land \ldots \land \sigma_k \rimp \psi_2$ is provable.  Since
$\psi_1 \Cond \sigma_j \in w$, $j = 1,\ldots, k$, it follows from
C0 that $\psi_1 \Cond (\sigma_1 \land \ldots \land \sigma_k) \in w$.
Now applying RC1, we get that $\psi_1 \Cond \psi_2 \in w$, as desired.

It follows that $\phi_0$ is satisfiable in $M$.

Now we have to show that $f$ satisfies all the properties in $\S$.
The arguments are straightforward. I consider three representative cases
here.
\begin{itemize}
\item[C1:] Suppose that $\mbox{C1} \in \C$.  We want to show that S1
holds. Thus, we must show that
$f(w,\phi) \subseteq \intension{\phi}_M$.  By definition, if $w' \in
f(w,\phi)$, then $g(w,\phi) \subseteq w'$.
If C1 holds, then $\phi \in g(w,\phi)$.  Thus, if $w' \in f(w,\phi)$,
then $\phi \in w'$.  By the Truth Lemma, $w' \in \intension{\phi}_M$.
\item[C4:] Suppose that $\mbox{C4} \in \C$ and $f(w,\phi) \ne
\emptyset$.  We want to show that $f(w,\phi)$ is a singleton.
Suppose that $w_1 \ne w_2$ are both in
$f(w,\phi)$.  Thus $g(w,\phi) \subseteq w_1 \inter w_2$.
There must be some formula $\psi$ such that $\psi \in w_1$
and $\neg \psi \in w_2$.
It follows from C4 that either $(M,w) \sat \phi \Cond \psi$ or
$(M,w) \sat \phi \Cond \neg \psi$.  In the former case, $\psi \in
g(w,\phi)$, while in the latter case, $\neg \psi \in g(w,\phi)$.  Either
way, it follows that $g(w,\phi) - w_1 \inter w_2 \ne \emptyset$, a
contradiction.
\item[C6:] Suppose  $\mbox{C6} \in \C$ and $f(w,\phi) \subseteq
\intension{\psi}_M$.  We want to show that $f(w,\phi \land \psi)
\subseteq f(w,\phi)$.  Suppose that $w' \in f(w,\phi \land \psi)$.  This
means that $g(w,\phi\land \psi) \subseteq w'$.  To show that $w' \in
f(w,\phi)$, we need to show that
$g(w,\phi) \subseteq w'$.  But if $\sigma \in g(w,\phi)$, then $\phi
\Cond \sigma \in w$.  By C6, it follows that $(\phi \land \psi) \Cond
\sigma \in w$, so $\sigma \in g(w,\phi \land \psi)$.  Since $g(w,\phi
\land \psi) \subseteq w'$, it follows that $\sigma \in w'$.  Thus,
$g(w,\phi) \subseteq w'$, as desired.
\end{itemize}

Finally, we must deal with the case that $\mbox{C8} \in \C$.  Let $w_0$
be a world in $M$ such that $\phi_0 \in w_0$.  (Recall that $\phi_0$ is
the formula  which we are trying to show is satisfiable.)  Let $W'$
consist of
all worlds in $W$ reachable from $w_0$.  It is almost immediate from the
definitions that if $w' \in W'$, then $f(w',\phi) \subseteq W'$ for all
formulas $\phi$.  Let $f'$ and $\pi'$ be the restrictions of $f$ and
$\pi$ to $W'$, respectively, and let $M(w_0) = (W',f',\pi')$.  The Truth
Lemma holds for $W'$; the proof above goes through without change.
Similarly the arguments that all the properties in $\S$ other than S8
hold in $M(w_0)$ go through without change.
Thus, $\phi_0$ is satisfiable in $M(w_0)$, and all the properties in $\S$
other than $\S$ also hold in $M(w_0)$.

To see that S8 also holds in $M(w_0)$,
suppose that $\intension{\phi}_{M(w_0)} \ne \emptyset$.  We want to
show that $f(w,\phi) \ne \emptyset$ for all $w \in W'$, or
equivalently, that $(M(w_0),w) \sat \Diamond \phi$.
Since all worlds in $W'$ are reachable from $w_0$, there must be two sequences of worlds,
$w_1, \ldots, w_k$ and $w_0',w_1', \ldots, w_m'$, and two sequences of
formulas, $\phi_1, \ldots, \phi_k$ and $\phi_1', \ldots, \phi_m'$ such
that $w_k = w$, $w_0' = w_0$, $(M(w_0),w_m') \sat \phi$,
$w_j$ is reachable from $w_{j-1}$ via $\phi_j$, for
$j=1, \ldots, k$, and $w_j'$ is reachable from $w_{j-1}'$ via $\phi_j'$
from $j= 1,\ldots, m$.  We must have $(M(w_0),w_0') \sat \Diamond \phi$, for
otherwise, by definition, we have $(M(w_0),w_0') \sat \Box \neg \phi$, and it
would follow by part (a) of Lemma~\ref{C8} that $(M(w_0),w_m') \sat \neg
\phi$, a contradiction.  Since $w_0' = w_0$, we have that $(M(w_0),w_0)
\sat \Diamond \phi$.  Now aplying part
(b) of Lemma~\ref{C8}, it follows that
$(M(w_0),w_k) \sat \Diamond \phi$.  Since $w_k = w$< we get the desired
result.  \eprf

\subsection{The Set-Theoretic Approach: Selection Functions}

In the set-theoretic approach, rather than having a syntactic operator
$\Cond$, we have a binary operator $\RCond$ on events (the superscript
$e$ stands for {\em event}); that is $\RCond : 2^\Omega \times 2^\Omega
\rightarrow 2^\Omega$.  For ease of exposition,
I write $H \RCond E$ instead of $\RCond(H,E)$.
Intuitively,
$\omega \in H \RCond E$ if, at world $\omega$, if $H$ were to hold, then
so would $E$.  Again, we can give semantics to $\RCond$ using selection
functions, but since we no longer have formulas, we replace the formula
that is the second argument in the syntactic case by a set of worlds
(which we can think of as the intension of a formula).  Thus, a
{\em (set-theoretic) selection
function}  maps $\Omega \times 2^\Omega$ to $2^\Omega$.

A {\em (set-theoretic) counterfactual structure\/} is then a pair $M =
(\Omega,\ftt)$, where $\Omega$ is a set of worlds and $\ftt$ is a
set-theoretic selection function.

We can then define the binary operator $\RCond: 2^\Omega \times
2^{\Omega}
\mapsto 2^{\Omega}$ in $M$ as follows:
\begin{equation}\label{RCond1}
H \RCond E  = \{\omega: \ftt(\omega,H) \subseteq E\}.
\end{equation}

We can define restrictions on $\ftt$ completely analogous to those
defined in the syntactic case.  These are listed
below, along with one
additional restriction, S9$'$.
\begin{itemize}
\item[S1$'$.] $\ftt(\omega,H) \subseteq H$
\item[S2$'$.] If $\ftt(\omega,H) \subseteq H'$ and
$\ftt(\omega,H') \subseteq H$, then $\ftt(\omega,H) =
\ftt(\omega,H')$
\item[S3$'$.] If $\omega \in H$, then $\ftt(\omega,H)= \{\omega\}$
\item[S4$'$.] $\ftt(\omega,H)$ is either empty or a singleton
\item[S5$'$.] $\ftt(\omega,H_1 \union H_2) \subseteq \ftt(\omega,H_1) \union
\ftt(\omega,H_2)$
\item[S6$'$.] If $\ftt(\omega, H) \subseteq E$ then $\ftt(\omega, H \inter E)
\subseteq \ftt(\omega,H)$%
\footnote{Again, the stronger version, with $\subseteq$ replaced by
equality, follows from S1$'$, S2$'$, and S5$'$, using
arguments almost identical to the earlier ones.  Similarly, a stronger
version of S7$'$ follows from S1$'$, S2$'$, and S5$'$.}
\item[S7$'$.] If $\ftt(\omega,H) \inter E \ne \emptyset$, then $\ftt(\omega, H
\inter E) \subseteq \ftt(\omega,H) \inter E$
\item[S8$'$.] If $H \ne \emptyset$ then $\ftt(\omega,H) \ne \emptyset$
\item[S9$'$.] If $\ftt(\omega,H) \subseteq E_1 \union E_2$, then there
exist $H_1, H_2$ such that $H_1 \union H_2 = H$, $\ftt(\omega,H_1)
\subseteq E_1$, and $\ftt(\omega,H_2) \subseteq E_2$
\end{itemize}

S9$'$ has no analogue in the syntactic case.
Since S5$'$ is
easily seen to be equivalent to ``if there exist
$H_1, H_2$ such that $H_1 \union H_2 = H$, $\ftt(\omega,H_1) \subseteq
E_1$, and $\ftt(\omega,H_2) \subseteq E_2$, then $\ftt(\omega,H)
\subseteq E_1 \union E_2$'', S9$'$ can be viewed as a converse to S5$'$.
It is also not hard to see that S9$'$ follows
from S1$'$ and S7$'$.  (Proof: if
$H \inter E_1 \ne \emptyset$ and $H \inter E_2 \ne \emptyset$, then
we can take $H_1 = H \inter E_1$  and $H_2 = H \inter (E_2
\union \neg H_1)$; if $H \inter E_1 = \emptyset$, we can take $H_1 =
\emptyset$, $H_2 = E$, and similarly if $H \inter E_2 = \emptyset$.)
As we shall see, S9$'$ arises (along with S1$'$, S2$'$, S5$'$, and
S6$'$) when the selection function
$\ftt$ is induced by an ordering on worlds.

Each of these restrictions corresponds to an axiom, completely analogous
to C0--C8.  Consider the following axioms:
\begin{itemize}
\item[C0$'$.]
$\inter_{j \in J} (H \RCond E_j) = H \RCond \inter_{j \in J}
E_J$ for any index set $J$ and events $E_j, j \in J$
\item[C1$'$.] $(H \RCond H) = \Omega$
\item[C2$'$.] $(H \RCond H') \inter (H' \RCond H) \inter (H \RCond E)
\subseteq (H' \RCond E)$
\item[C3$'$.] $H \inter (H \RCond E) = H \inter E$
\item[C4$'$.] $(H \RCond E) \union (H \RCond \neg E) = \Omega$
\item[C5$'$.] $(H_1 \RCond E) \inter
     (H_2 \RCond E) \subseteq (H_1 \union H_2) \RCond E$
\item[C6$'$.] $(H \RCond E_1) \inter (H \RCond E_2) \subseteq
(H \inter E_1) \RCond E_2$
\item[C7$'$.] $\neg (H \RCond \neg E_1) \inter (H \RCond E_2)
           \subseteq (H \inter E_1) \RCond E_2$
\item[C8$'$.] $H \RCond \emptyset = \emptyset$ if $H \ne \emptyset$
\item[C9$'$.] $H \RCond (E_1 \union E_2) \subseteq \union_{\{H_1, H_2:
H_1\union H_2 = H\}} ((H_1 \RCond E_1) \inter (H_2 \RCond E_2))$
\end{itemize}

We can now state
and prove
the semantic soundness and completeness result.
\thmm\label{char1} Let $\S$ be a (possibly empty) subset of
$\{\mbox{S1$'$, \ldots, S9$'$}\}$,
let $\C$ be the corresponding subset of $\{\mbox{C1$'$,\ldots, C9$'$}\}$,
and let $\Omega$ be a set of worlds.  If $\ftt$ is a set-theoretic
selection function on
$\Omega$ that satisfies the conditions in $\S$ and $\RCond$ is defined
in $(\Omega,\ftt)$ by (\ref{RCond1}), then $\RCond$
satisfies C0$'$ and the axioms in $\C$.  Conversely, if the function
$\leadsto: 2^\Omega \times 2^\Omega \mapsto 2^\Omega$
satisfies C0$'$ and the axioms in $\C$, then there is a selection
function $\ftt$ on $\Omega$ satisfying $\S$ such that $\leadsto$ is the
counterfactual operator $\RCond$ in $(\Omega,\ftt)$. \ethm

\prf It is easy to check that if $\ftt$ satisfies the conditions in $\S$, then
$\RCond$ satisfies C0$'$ and all the conditions in $\C$.  For the second
half, given a function
$\leadsto: 2^\Omega \times 2^\Omega \mapsto 2^\Omega$
that satisfies C0$'$ and the axioms in $\C$,
define $\ftt(\omega,H) = \inter \{E:
\omega \in H \leadsto E\}$.  I leave it to the reader to check that
$\leadsto$ is the counterfactual operator $\RCond$ in $(\Omega,\ftt)$
and satisfies $\S$.
\eprf

Again, a number of points are worth making.  First,
observe how C8$'$ captures S8$'$ far more directly than C8
captures S8.  We did not have to struggle to find an axiom corresponding
to this condition.  Next,
observe that the completeness proof proceeds in somewhat the
same spirit as the syntactic completeness proof, but avoids the
construction of a canonical model.  It works whichever model we start
with.
An analogue to Corollary~\ref{completecor} can also be proved, showing
that, again, we are entitled to view Theorem~\ref{char1} as a soundness
and completeness result. Finally,
as I now show, we need the full strength of C0$'$ for completeness.

Consider the obvious finitary analogue of C0$'$:
\begin{itemize}
\item[C10$'$.] $(H \RCond E_1) \inter (H \RCond E_2) = H \RCond (E_1
\inter E_2)$
\end{itemize}
I actually give two examples showing that we cannot in general
replace C0$'$ with C10$'$.   The first is a simple example
that satisfies C1$'$--C3$'$, C5$'$--C10$'$, but not C0$'$ (or C4$'$).
Then I give a somewhat more sophisticated example that satisfies
all of C1$'$--C10$'$, but not C0$'$.  Since $\RCond$ must satisfy C0$'$,
this shows that C10$'$ does not suffice, even in the presence of all the
other properties.

\xam\label{simplexam}
Let $\Omega = \{1,2,3, \ldots\}$.  Define $\leadsto$ so that
if
$\omega \in H$, then $\omega \in H \leadsto E$ iff $w \in E$; if $\omega
\notin H$, then $w \in H \leadsto E$ iff
(a) $H \inter \neg E = \emptyset$ or (b) $H$ is infinite and $H \inter
\neg E$ is finite.
I leave to the reader the somewhat tedious (but straightforward) task of
checking the $\leadsto$ satisfies C1$'$--C3$'$, C5$'$--C10$'$.  It does
not satisfy C0$'$, since if $E_j = \neg\{j\}$, then $E_1 \leadsto
E_1 = \Omega$ and $E_1 \leadsto E_j = E_j$ 
for $j = 2, 2, 3, \ldots$.  But
$\inter_j E_j = \emptyset$, $E_1 \leadsto \emptyset = \emptyset$, and
$\inter_j (E_1 \leadsto E_j) = \{1\}$.  Thus, C0$'$ does not hold.
Note that $\leadsto$ also does not satisfy C4$'$
since, for example,
$(\{1,2\} \leadsto \{1\}) \union (\{1,2\} \leadsto \neg\{1\}) = \{1,2\}
\ne \Omega$.
\exam

\xam
For this example, we need to review some material on filters and
ultrafilters \cite{BellSlomson}.  A {\em filter\/} on $\Omega$ is a
nonempty set $\U$
of subsets of $\Omega$ that is closed under supersets (\ie if $E \in \U$
and $E \subset E'$, then $E' \in \U$) and finite intersections, and does
not contain the empty set.  An {\em ultrafilter\/} is a maximal filter,
that
is, a filter that is not a subset of any other filter.  A {\em principal
ultrafilter\/} is an ultrafilter which consists of all the supersets of a
particular element of $\Omega$.  (It is easy to check that this is
indeed an ultrafilter.) A {\em nonprincipal ultrafilter\/} is an
ultrafilter that is not a principal ultrafilter.   Note that if $\U$ is
an ultrafilter, then for any set $H$, either $H$ or $\neg H$ must be in
$\U$.  From this it follows that if  $H \in \U$, then for any set
$E$, we must have
that one of $H \inter E$ or $H \inter \neg E$ is in $\U$.  (Proof: If
neither is in $\U$, then $H$, $\neg H \union E$, and $\neg H \union \neg
E$ are all in $\U$.  Since $\U$ is closed under finite intersections, it
follows that $\emptyset \in \U$.  This contradicts the fact that $\U$ is
a filter.)

Let $\U$ be a nonprincipal ultrafilter on $\Omega = \{1,2,3, \ldots\}$.
(Nonprincipal
ultrafilters on $\Omega$ can be shown to exist using Zorn's Lemma.)
Define
$\leadsto'$ as follows: $\omega \in H\leadsto' E$ if and only if
the following conditions hold:
\begin{itemize}
\item[(a)] if $\omega\in H$, then $\omega \in H \inter E$,
\item[(b)] if $\omega \notin H$ and $H \in \U$, then $H \inter E \in \U$,
\item[(c)] if $\omega \notin H$ and $H \notin \U$, then $H = \emptyset$
or $\min(H \inter E) = \min(H)$ (where $\min(F)$ denotes the minimal
element of $F$).
\end{itemize}
Again, it is straightforward (but tedious) to show that $\leadsto'$
satisfies C1$'$--C10$'$.  For example, to see that C4$'$ holds,
we must show that, for all $E$, we have
$(H \leadsto' E) \union (H \leadsto' \neg E) = \Omega$.   We do
a case-by-case analysis.  If $\omega \in H$, then, by clause (a), then
if $\omega \in E$, we also have $\omega \in H \leadsto' E$, by
definition; otherwise, $\omega \in H \leadsto' \neg E$.
If $\omega \notin H$ and $H \in \U$, then
one of $H \inter E$ or $H \inter
\neg E$ must be in $\U$, since otherwise, since $\U$ is an ultrafilter,
we would have both $\neg H \union E \in \U$ and
$\neg H \union \neg E \in \U$, from which
it follows that $\neg H \in U$,
contradicting the assumption that $H \in \U$.
Finally, if $\omega \notin H$ and $H \notin \U$,
then either $\min(H) = \min(H \inter E)$ or $\min(H) = \min(H \inter
\neg E)$.
In all cases, we have $\omega \in (H \leadsto' E) \union (H \leadsto'
\neg E)$.

To see that $\leadsto'$ does not satisfy C0$'$, note that a
nonprincipal ultrafilter can contain only
infinite sets, and so must contain all cofinite sets.  Taking the
sets $E_i$ as in Example~\ref{simplexam}, it follows that $E_i$ and $E_i
\inter E_j$ are both in $\U$, for all $i, j$ (since these are all
cofinite sets).  It then follows from the definition that $E_1 \leadsto'
E_1 = \Omega$ and $E_1 \leadsto' E_j = E_j$.  Thus, we get a violation
of C0$'$ just as in Example~\ref{simplexam}.
\exam

\subsection{Preferential Orders}\label{ordering}
I have described $f(\omega,\phi)$ as the ``closest'' worlds to $\omega$
satisfying $\phi$.  This suggests that there is an underlying ordering on
worlds.  Lewis \citeyear{Lewis73} made this intuition explicit as follows.
A {\em preferential frame\/} is a pair $(\Omega,R)$, where $R$ is
a ternary {\em preferential relation\/} on a set $\Omega$ of possible
worlds.   For technical reasons that will be
discussed at the end of this subsection, I assume that $\Omega$ is
finite in this subsection. We typically write $\omega' \preceq^{\omega}
\omega''$ rather than $R(\omega,\omega',\omega'')$.
This should be thought of as saying that $\omega'$ is at least as close
to $\omega$ as
$\omega''$; thus, $\preceq^\omega$ represents the ``at least as close to
$\omega$'' relation.
Let $\Omega_w = \{w': \exists w''(w' \preceq^w
w'')\}$.
We can think of $\Omega_w$ as the domain of $\preceq^w$.
Intuitively, the worlds not in $\Omega_w$ are so far away from $w$ that
they cannot even be discussed.
As would be expected from the
intuition, we require that $\preceq^\omega$ be a partial preorder on
$\Omega_w$, that is, a reflexive, transitive relation.%
\footnote{Note that $\preceq^\omega$ is not necessarily anti-symmetric.
That is why it is a preorder, not an order.  Nor is it necessarily
total; totality is forced by condition P2, which will be
defined shortly.}
We define the relation $\prec^\omega$ by taking $\omega' \prec^\omega
\omega''$ if $\omega' \preceq^\omega \omega''$ and not($\omega''
\preceq^\omega \omega'$).

Given our intuitions regarding closeness, the following
requirements seem reasonable:
\begin{itemize}
\item[P1.] $\omega \in \Omega_w$ and is the minimal element with respect
to $\preceq^\omega$ (so that $\omega$ is closer to itself than
any other element): formally, for all $\omega'
\in \Omega_w$, we have $\omega \preceq^\omega \omega$ and
$\omega \prec^\omega \omega'$.
\item[P2.] $\preceq^\omega$ is a total preorder on $\Omega_w$: that is,
for all $\omega', \omega'' \in \Omega_w$,
either $\omega' \preceq^\omega \omega''$ or
$\omega'' \preceq^\omega \omega'$.
\item[P3.] $\preceq^\omega$ is a linear order on $\Omega_w$: that is,
for all $\omega', \omega'' \in \Omega_w$, if $\omega' \ne \omega''$,
then either $\omega' \prec^\omega \omega'$ or $\omega'' \prec^\omega
\omega'$.
\item[P4.] $\Omega_w = \Omega$ for all $w \in \Omega$
\end{itemize}

In a preferential frame, we can define a (set-theoretic) selection
function $\ftt_\preceq$ such that $\ftt_\preceq(\omega,H)$ are the worlds
closest to $\omega$, according to $\preceq^\omega$, that are in $H$.
Formally, we have
$$\ftt_\preceq(\omega,H) = \{\omega' \in H \inter W_w: \mbox{if } \omega''
\prec^\omega \omega' \mbox{ then } \omega'' \notin H\}.$$
This gives us a way of defining a binary operator $\RCond$ in
preferential
frames, by an immediate appeal to Definition~(\ref{RCond1}).

We can also define a syntactic version (which is actually what Lewis
did).  A {\em preferential structure\/} is a tuple $M = (\Omega,R,\pi)$,
where $(\Omega,R)$ is a preferential frame and $\pi$ is an
interpretation.  Roughly speaking, we would now like to define
a syntactic selection function $f_\preceq$ in $M$ by taking
\begin{equation}\label{RCond2}
f_\preceq(\omega,\phi) = \{\omega' \in \intension{\phi}_M \inter W_w: \mbox{if }
\omega'' \prec^\omega \omega' \mbox{ then } \omega'' \notin
\intension{\phi}_M\},
\end{equation}
and then use this selection function to give semantics to conditional
formulas.  The only problem is that we have not yet defined
$\intension{\phi}_M$.  The formal definition does not use the
selection function directly.  Nevertheless, it is easy to check that the
formal definition is consistent with this intuition.

Formally, the definition of $\sat$ in preferential structures is the
same as that in counterfactual structures, except for the clause for
$\Cond$, which is
\begin{description}
\item $(M,w) \sat \phi \Cond \psi$ if
$\ftt_\preceq(w, \intension{\phi}_M)
\subseteq \intension{\psi}_M$
\end{description}
This is well-defined, since the inductive definition of $\sat$
guarantees that the sets $\intension{\phi}_M$ and $\intension{\psi}_M$
have already been defined.   This definition makes precise the intuition
that $\phi \Cond
\psi$ holds at world $w$ if the $\phi$-worlds closest to $w$ (according
to the ordering $\preceq^w$ used at $w$) all satisfy $\psi$.

The following syntactic completeness result is well known.  It says that
in preferential structures, $\Cond$ satisfies C0, C1, C2, C5, C6, and
RC1; moreover, P1, P2, P3, and P4 give us C3, C7, C4, and C8,
respectively.
\thmm\label{char2syn} \cite{Burgess81,FrH3,Lewis73}
Let $\P$ be any (possibly empty) subset of
$\{$P1,P2,P3,P4$\}$
let $\C$ be the corresponding subset of $\{$C3,C7,C4,C8$\}$, and
let $\Omega$ be a finite set of worlds.  Let $\M^\P$ be the class of
preferential structures where the ternary relation $R$
satisfies
the conditions in $\P$.  Then $\C \union \{\mbox{Prop, C0, C1, C2,
C5, C6, MP, RC1}\}$ is a
sound and complete axiomatization for the language $\LPC$ for the class
of preferential structures in $\M^\P$.%
\footnote{Note that since C1, C5, C6, RC1, and LLE all hold (recall that
LLE follows from C1, C2, and RC1), we can replace C8 by the simpler
$(\Box \phi \rimp (\Box \Box \phi \land \phi)) \land
(\Diamond \phi \rimp \Box \Diamond \phi)$.}
\ethm

What about set-theoretic completeness?  Not surprisingly in the light
of the previous theorem, it turns out that $f_\preceq$ satisfies S1$'$,
S2$'$, S5$'$, S6$'$; moreover, P1, P2, P3, and P4 correspond to
S3$'$, S7$'$, S4$'$, and S8$'$, respectively.  The interesting thing is
that we also get S9$'$.

\lemm\label{fttchar} Let $\P$ be a subset of $\{\mbox{P1,P2,P3,P4}\}$,
and let $\S$
be the corresponding subset of $\{\mbox{S4$'$,S7$'$,S4$'$,S8$'$}\}$.   If
$(\Omega,R)$
is a preferential frame satisfying the properties in $\P$, then
$\ftt_\preceq$ satisfies S1$'$, S2$'$,
S5$'$, S6$'$, S9$'$ and all the properties in $\S$. \elem

\prf Proving that $\ftt_\preceq$ satisfies all the properties is
straightforward.
I consider only S9$'$ here.  To see that
$f_\preceq$ satisfies S9$'$, suppose
$\ftt_\preceq(\omega,H) \subseteq E_1 \union E_2$.
Let $E_i' = E_i \inter f(w,H)$, for $j = 1,2$.
Let $E_j^H$
consist of all the elements in $H$ that are at least as far from
$\omega$ as some element in $E_j$.
That is,
$E_j^H = \{\omega' \in H: \exists \omega'' \in E_j'( \omega''
\preceq^\omega \omega')\}$
 Then define
$H_1 =  (E_1^H - (E_2' - E_1')) \union (H - W_\omega)$ and
$H_2 =  (E_2^H - (E_1' - E_2')) \union (H - W_\omega)$.  It is not hard
to show that
$H = H_1 \union H_2$.  (Proof: By construction, $A^H \subseteq H$,
so we must have $H_1 \union
H_2 \subseteq H$.  For the opposite containment, note that
if $v \in H - W_\omega$, then the construction guarantees that
$v \in H_1 \inter H_2$.  Also note that $E_j' \inter W_\omega \subseteq
E_j^H$, since if $v \in E_j'$ then $v \in E_j \inter H$ and if $v \in
W_\omega$, then $v \preceq^w v$.  Thus, $E_j' \subseteq H_j$ for $j
=1,2$.  Finally, if $v \in H \inter W_\omega - (E_1' \union E_2')$,
then choose
$v' \in \ftt_{\preceq}(\omega,H)$ such that $v' \preceq^w v$.  Such a $v'$
must exist by definition, $v' \in H$ by construction, and $v' \in E_1
\union E_2$ since $\ftt_{\preceq}(\omega,H) \subseteq E_1 \union E_2$.
If $v' \in E_j$, then we must have $v \in E_j^H$ and hence $v \in H_j$.)

It remains to show that
$\ftt_\preceq(\omega,H_j) \subseteq E_j$, $j = 1,2$.  I consider the
case that $j=1$.  (The proof for $j=2$ is almost identical.)
Suppose that $v \in \ftt_\preceq(\omega,H_1)$.
Thus, $v \in E_1^H - (E_2' - E_1')$, since
$\ftt_\preceq(\omega,H_1) \subseteq \Omega_w$.  Since $v \in E_1^H$,
there exists some $v' \in E_1'$ such that $v' \preceq^\omega v$.  Since,
as we observed above, $E_1' \subseteq H_1$, we have $v' \in H_1$.  Thus
we  cannot have $v' \prec^\omega v$, for otherwise $v
\notin \ftt_\preceq(\omega,H_j)$.  Thus, $v \preceq^\omega v'$.  It
follows that $v \in \ftt_\preceq(\omega,H)$, for otherwise there would
be some $v'' \in E_1' \union E_2'$ such that $v'' \prec^\omega v$.  This
would mean that $v'' \prec^\omega v'$, contradicting the fact that $v'
\in E_1'$.  Thus, it follows that $v \in E_1' \union E_2'$.  Since $v
\in H_1$, we cannot have $v \in E_2' - E_1'$.  Hence $v \in E_1'$, as
desired.
\eprf

We then get the following set-theoretic soundness and completeness
result.  Note that we can use C10$'$ instead of C0$'$, since we are
restricting to finite sets of worlds.
\thmm\label{char2}  Let $\P$ be any (possibly empty) subset of
$\{$P1,P2,P3,P4$\}$,
let $\C$ be the corresponding subset of $\{\mbox{C3$'$,C7$'$,C4$'$,C8$'$}
\}$, and let $\Omega$ be a finite set of worlds.  If $R$ is a ternary
relation on $\Omega$ that satisfies
the conditions in $\P$ then $\RCond$
satisfies C1$'$, C2$'$, C5$'$, C6$'$, C9$'$, C10$'$, and all the axioms
in $\C$. Conversely, if $\leadsto:
2^\Omega \times 2^\Omega \mapsto 2^\Omega$
and satisfies C1$'$, C2$'$, C5$'$, C6$'$, C9$'$, C10$'$, and the axioms
in $\C$, then there
is a ternary relation $R$ on $\Omega$ satisfying the conditions in $\P$
such that
$\leadsto$ is the counterfactual operator $\RCond$ in $(\Omega,R)$.
\ethm

\prf  The first half (soundness) follow immediately from
Lemma~\ref{fttchar} and
Theorem~\ref{char1}.  For the second half, suppose that $\leadsto$
satisfies
C1$'$, C2$'$, C5$'$, C6$'$, C9$'$, C10$'$, and the axioms
in $\C$.  First assume that C7$' \notin \C$.
Define $\omega' \preceq^\omega \omega''$ iff $\omega \in
\{\omega',\omega''\} \leadsto \{\omega'\}$ and $\omega \notin
\{\omega',\omega''\} \leadsto \emptyset$.
(This way of defining the
ordering is essentially due to \cite{KLM}; a slightly different ordering
for the case that C7$'$ is in $\C$ is described at the end of the
proof.)
We must show that $\preceq^\omega$
is a partial order on $W_\omega$ and that the
operator $\RCond$ determined by this relation agrees with $\leadsto$.
The following lemma will prove useful:

\lemm\label{W_w} $W_\omega = \{\omega':
\omega \notin \{\omega'\} \leadsto \emptyset\}$. \elem

\prf  Let $W' = \{\omega':
\omega \notin \{\omega'\} \leadsto \emptyset\}$.
If $\omega' \in W'$, then we have $\omega' \preceq^\omega \omega'$
(since, by C1$'$, we must have $\omega\in \{\omega'\} \leadsto
\{\omega'\}$) and thus $\omega' \in \Omega_\omega$.  Conversely, if
$\omega' \in \Omega_\omega$, then there must be some $\omega''$ such
that $\omega'  \preceq^\omega \omega''$.
Thus,
\begin{equation}\label{new1}
\omega \in \{\omega', \omega''\} \leadsto \{\omega'\}
\end{equation}
and
\begin{equation}\label{new2}
\omega \notin \{\omega',\omega''\} \leadsto \emptyset.
\end{equation}
Suppose, by
way of contradiction,  that
\begin{equation}\label{new2.5}
\omega \in \{\omega'\} \leadsto \emptyset.
\end{equation}
From C0$'$ and (\ref{new2.5}), it follows that
\begin{equation}\label{new3}
\omega \in \{\omega'\} \leadsto \{\omega''\}.
\end{equation}
By C1$'$, we have
\begin{equation}\label{new4}
\omega \in \{\omega''\} \leadsto \{\omega''\}.
\end{equation}
By C5$'$ applied to (\ref{new3}) and (\ref{new4}), we have
\begin{equation}\label{new5}
\omega \in \{\omega',\omega''\} \leadsto \{\omega''\}.
\end{equation}
Finally, from C10$'$ applied to (\ref{new1}) and (\ref{new5}), we get
$\omega \in \{\omega',\omega''\} \leadsto \emptyset$.  But this
contradicts (\ref{new2}). \eprf

It follows immediately from C1$'$ and Lemma~\ref{W_w} that
$\preceq^\omega$ is reflexive on $\Omega_\omega$.  To show that it is
transitive, it is easy to see that
it suffices to show that if $E_1, E_2, E_3$ are disjoint sets, then
$$((E_1 \union E_2) \leadsto E_1) \inter ((E_2 \union E_3) \leadsto E_2)
\subseteq ((E_1 \union E_3) \leadsto E_1).$$
(Transitivity follows easily from the special case that $E_j =
\{\omega_j\}$, $j = 1, 2, 3$, where $\omega_1$, $\omega_2$, and
$\omega_3$ are different worlds.)

For the proof, it is useful to have two preliminary lemmas.
\lemm\label{prelim1}
If $E \subseteq E'$ then $H \leadsto E \subseteq H \leadsto E'$.
\elem
\prf If $E \subseteq E'$, then by C10$'$ we have
$$(H \leadsto E) \inter (H
\leadsto E') = H \leadsto (E \inter E') = H \leadsto E. \ \ \ \ $$ 
\eprf

\lemm\label{prelim2}
$(H_1 \leadsto E_1) \inter (H_2 \leadsto E_2) \subseteq (H_1 \union H_2)
\leadsto (E_1 \union E_2)$.  \elem

\prf Using Lemma~\ref{prelim1} and C5$'$, we have
$$(H_1 \leadsto E_1) \inter (H_2 \leadsto E_2) \subseteq (H_1 \leadsto (E_1
\union E_2)) \inter (H_2 \leadsto (E_1 \union E_2)) \subseteq
(H_1 \union H_2) \leadsto (E_1 \union E_2).$$
\eprf

From Lemma~\ref{prelim2}, we have that
\begin{equation}\label{eq3}
((E_1 \union E_2) \leadsto E_1) \inter ((E_2 \union E_3) \leadsto E_2)
\subseteq
(E_1 \union E_2 \union E_3) \leadsto (E_1 \union E_2)
\end{equation}
Applying C1$'$ and Lemma~\ref{prelim2}, we also get
\begin{equation}\label{eq4}
(E_1 \union E_2) \leadsto E_1 =
((E_1 \union E_2) \leadsto E_1) \inter (E_3 \leadsto E_3) \subseteq
(E_1 \union E_2 \union E_3) \leadsto (E_1 \union E_3)
\end{equation}
Finally, using (\ref{eq3}), (\ref{eq4}), C10$'$, and C6$'$, we get
$$\begin{array}{lll}
&((E_1 \union E_2) \leadsto E_1) \inter ((E_2 \union E_3) \leadsto E_2)\\
\subseteq
&((E_1 \union E_2 \union E_3) \leadsto (E_1 \union E_2)) \inter
((E_1 \union E_2 \union E_3) \leadsto (E_1 \union E_3))\\
= &((E_1 \union E_2 \union E_3) \leadsto E_1) \inter
((E_1 \union E_2 \union E_3) \leadsto (E_1 \union E_3))\\
\subseteq &(E_1 \union  E_3) \leadsto E_1.
\end{array}$$
Thus $\preceq^\omega$ is transitive, as desired.

Next, we must show that C3$'$, C4$'$, and C8$'$ give us P1, P2, and P4,
respectively.  (Recall that I have deferred the case of C7$'$.)

Suppose that $\leadsto$ satisfies
C3$'$.  Then $\{\omega\} \inter (\{\omega\} \leadsto \emptyset) =
\emptyset$,
so $\omega \notin \{\omega\} \leadsto \emptyset$; thus, $\omega \in
W_\omega$ by Lemma~\ref{W_w}.  Moreover, since $\{\omega, \omega'\}
\inter (\{\omega,\omega'\} \leadsto
\{\omega\})  = \{\omega\}$, we have $\omega \preceq^\omega \omega'$
for all $\omega' \in W_\omega$, showing that P1 holds.

If $\leadsto$ satisfies C4$'$, then (using C1$'$ and C10$'$), we have
$$\begin{array}{ll}
W &= \{\omega', \omega''\} \leadsto \{\omega'\} \union
\{\omega', \omega''\} \leadsto (W - \{\omega'\})\\
  &= \{\omega', \omega''\} \leadsto \{\omega'\} \union
(\{\omega', \omega''\} \leadsto (W - \{\omega'\}) \inter
\{\omega', \omega''\} \leadsto \{\omega', \omega''\})\\
  &= \{\omega', \omega''\} \leadsto \{\omega'\} \union
\{\omega', \omega''\} \leadsto \{\omega''\}.
\end{array}
$$
It easily follows that for all $\omega', \omega'' \in W_w$, we have
either $\omega' \preceq^\omega \omega''$ or
$\omega'' \preceq^\omega \omega'$.  Thus, P2 holds.

Finally, it is immediate from the definitions that if $\leadsto$
satisfies C8$'$, then $W_\omega = W$.

It remains to check that $\RCond$ and $\leadsto$ agree.
We prove that
\begin{equation}\label{agree}
\mbox{$H \RCond E$ iff $H \leadsto E$.}
\end{equation}

Note that C1$'$, C2$'$, C5$'$, C6$'$, and C10$'$ hold for both $\RCond$
and $\leadsto$.
For $\RCond$, this follows from the first half of the theorem; for
$\leadsto$, it follows by assumption.  These are the only properties used
in the proof.  Note that this means that Lemmas~\ref{prelim1}
and~\ref{prelim2} apply to both $\RCond$ and $\leadsto$.

Using C10$'$ and C1$'$, we have that $H \RCond E = H
\RCond (H \inter E)$, and similarly for $\leadsto$; thus, we can assume
without loss of generality that $E \subseteq H$.
We proceed by induction on $|E|$.
(Here we are making heavy use of the fact that $\Omega$ is finite.)
If $E = \emptyset$, then if $H = \emptyset$, the result follows from
C1$'$.  If $H \ne \emptyset$, then for each $\omega' \in H$, we have
$$H \leadsto \emptyset = ((H \leadsto \emptyset) \inter (H
\leadsto \{\omega'\}))
\subseteq \{\omega'\} \leadsto \emptyset.$$
Thus, we have
$$H \leadsto \emptyset \subseteq \inter_{\omega' \in H} \{\omega'\}
\leadsto \emptyset.$$
By C5$'$, we actually have
$$H \leadsto \emptyset = \inter_{\omega' \in H} \{\omega'\}
\leadsto \emptyset.$$
An identical argument works if we replace $\leadsto$ by $\RCond$.
Thus, it suffices to show that $\{\omega'\} \leadsto \emptyset =
\{\omega'\} \RCond \emptyset$.  But, by Lemma~\ref{W_w},
$\omega \in \{\omega'\} \leadsto \emptyset$ iff $\omega' \notin
W_\omega$.  It is also immediate from the definitions that $\omega'
\notin W_\omega$  iff $\omega \in \{\omega'\} \RCond \emptyset$.
Thus, $\{\omega'\} \leadsto \emptyset =
\{\omega'\} \RCond \emptyset$, as desired.

If $|E|=1$, suppose $E = \{\omega'\}$.
We proceed by a subinduction on $|H|$.
If $|H| =1$, then the result is immediate from C1$'$, since
$E \subseteq H$.  If $|H|=2$, let $H = \{\omega',\omega''\}$.
If $H \leadsto E$, there are a number of cases to
consider.  First suppose that $\omega',\omega'' \in W_\omega$.  Then
$\omega \in \{\omega',\omega''\} \RCond \{\omega'\}$ iff
$\omega' \prec^{\omega} \omega''$, which implies
$\omega \in \{\omega',\omega''\} \leadsto \{\omega'\}$ by definition.
If $\omega' \notin W_\omega$ then,by Lemma~\ref{W_w}, we have  $\omega
\in \{\omega'\} \leadsto \emptyset$.  By C1$'$, we have $\omega \in
\{\omega''\} \leadsto \{\omega''\}$.  Now by Lemma~\ref{prelim2}, we have
$\omega \in \{\omega',\omega''\} \leadsto \{\omega''\}$.  By C10$'$, it
follows that $\omega \in \{\omega',\omega''\} \leadsto \emptyset$.  By
the induction hypothesis, we have $\omega \in \{\omega',\omega''\}
\RCond \emptyset$, and by Lemma~\ref{prelim1}, we have $\omega
\in \{\omega',\omega''\} \RCond \{\omega'\}$, as desired.
Finally, if $\omega'' \notin W_\omega$, then $\omega \in \{\omega''\}
\leadsto \emptyset$.  By the induction hypothesis, we have $\omega \in
\{\omega''\} \RCond \emptyset$, and by C1$'$, we have $\omega \in
\{\omega'\} \RCond \{\omega'\}$.  Now by Lemma~\ref{prelim2}, we have
$\omega \in \{\omega',\omega''\} \RCond \{\omega'\}$, as desired.

For the converse, note that if
$\omega \in \{\omega',\omega''\} \RCond \{\omega'\}$, then either
$\omega' \prec^{\omega} \omega''$ or $\omega'' \notin W_\omega$.
If $\omega' \prec^{\omega} \omega''$, then we have
$\omega \in \{\omega',\omega''\} \leadsto \{\omega'\}$ by definition,
while if $\omega'' \notin W_\omega$, then by Lemma~\ref{W_w}, we have
$\omega \in \{\omega''\}
\leadsto \emptyset$.  Since, by C1$'$, we also have $\omega \in
\{\omega'\} \leadsto \{\omega'\}$, the desired result follows from
Lemma~\ref{prelim2}.

To complete the subinduction,
suppose $|H| > 2$.
If $\omega \in H \leadsto \{\omega'\}$, then by
Lemma~\ref{prelim1},
we also have $\omega \in H \leadsto \{\omega',\omega''\}$ for all
$\omega'' \in H$.
Thus, by C6$'$, $\omega \in \{\omega',\omega''\} \leadsto \{\omega'\}$
for all $\omega''
\in H$.  By the induction hypothesis, we have
$\omega \in \{\omega',\omega''\}
\RCond \{\omega'\}$ for all $\omega'' \in H$.  By C5$'$, we have $\omega
\in H \RCond \{\omega'\}$.  A symmetric argument works for the converse
(replacing the roles of $\RCond$ and $\leadsto$).

Finally, suppose $|E| > 1$.  Choose some $\omega' \in E$. By C9$'$,
we have that if $\omega \in H \RCond E$, then $\omega \in (H_1 \leadsto
(E - \{\omega'\})) \inter
(H_2 \RCond \{\omega'\})$ for some $H_1, H_2$ such that $H_1 \union H_2 =
H$.  By the induction hypothesis, $H_1 \RCond E-\{\omega'\} = H_1
\leadsto E-\{\omega'\}$ and $H_2 \RCond \{\omega'\} = H_2 \leadsto
\{\omega'\}$.  By Lemma~\ref{prelim2}, $H_1 \leadsto E - \{\omega'\}
\inter H_2 \leadsto \{\omega'\} \subseteq H \leadsto E$.  Thus,
$\omega \in H \leadsto E$.  The opposite
containment is obtained by a symmetric argument.

Now we must deal with the case that $\mbox{C7$'$} \in \C$.  The argument
is similar in spirit to that given in \cite{FrH5Full}.  In this case,
$\preceq^\omega$ is not necessarily a total order.  However, we can show
that $\prec^\omega$ is {\em modular}: if $\omega_1 \prec^\omega
\omega_2$, then for all $\omega_3 \in W_\omega$, either $\omega_3
\prec^\omega
\omega_2$ or $\omega_1 \prec^\omega \omega_3$. To see this, suppose
$\omega_1 \prec^\omega \omega_2$ and it is not the case that $\omega_3
\prec^\omega \omega_2$.  Then
\begin{equation}\label{eq5}
\omega \in \{\omega_1, \omega_2\} \leadsto \{\omega_1\}
\end{equation}
and
\begin{equation}\label{eq6}
\omega \notin \{\omega_2,\omega_3\} \leadsto \{\omega_3\}.
\end{equation}

Since $\omega \in \{\omega_3\} \leadsto \{\omega_3\}$, from (\ref{eq5})
and Lemma~\ref{prelim2} it follows that
\begin{equation}\label{eq7}
\omega \in \{\omega_1, \omega_2,\omega_3\} \leadsto
\{\omega_1,\omega_3\}.
\end{equation}
We also must have
\begin{equation}\label{eq8}
\omega \in \{\omega_1, \omega_2,\omega_3\} \leadsto \{\omega_1\},
\end{equation}
for otherwise, we have $\omega \in \neg
(\{\omega_1, \omega_2,\omega_3\} \leadsto \neg \neg \{\omega_1\})$, so
by C7$'$ and (\ref{eq7}), we would have
\begin{equation}\label{eq9}
\omega \in \{\omega_2, \omega_3\} \leadsto \{\omega_1,\omega_3\}.
\end{equation}
Since we also have
$\omega \in \{\omega_2, \omega_3\} \leadsto \{\omega_2,\omega_3\}$, an
application of C10$'$ gives us
$\omega \in \{\omega_2,\omega_3\} \RCond
\{\omega_3\}$, contradicting~(\ref{eq6}).
From~(\ref{eq7}), (\ref{eq8}), and C6$'$, we immediately get
\begin{equation}\label{eq10}
\omega \in \{\omega_1,\omega_3\} \RCond \{\omega_1\}.
\end{equation}

We also cannot have $\omega \in \{\omega_1,\omega_3\} \RCond
\{\omega_3\}$, for then by C10$'$ and (\ref{eq10}), we would have
$\omega \in
\{\omega_1,\omega_3\} \RCond \emptyset$.  From Lemma~\ref{prelim1}, we
then would get $\omega \in \{\omega_1,\omega_3\}\RCond \{\omega_3\}$,
and from C6$'$, that $\omega \in \{\omega_3\} \RCond \emptyset$,
contradicting the assumption that $\omega_3 \in W_{\omega}$.  We
can therefore conclude, using (\ref{eq10}), that  $\omega_1 \prec^\omega
\omega_3$, giving us the desired modular order.

Once we have a modular order, we can easily define a total order from
it.  Define $\omega' \le^\omega \omega''$ either if $\omega'
\prec^\omega \omega''$ or neither $\omega' \prec^\omega \omega''$
nor $\omega'' \prec^\omega \omega'$ hold.  It is a standard result
(and not hard to show) that $\le^\omega$ is a total order if
$\prec^\omega$ is modular.  (See
\cite[Lemma 2.6]{Hal17} for a proof.)  Moreover, it is easy to
see $\omega' <^\omega \omega''$ iff $\omega' \prec^\omega \omega''$.
Thus, $f_\le = \best$, so that if we use $\le^\omega$ to define the
ternary relation, our previous argument shows that $\RCond$ and
$\leadsto$ still agree.  This completes the proof of Theorem~\ref{char2}.
Although the proof is certainly nontrivial, it is still
significantly simpler than the corresponding syntactic proof (see,
for example, \cite{Burgess81,FrH3}).
\eprf

I conclude with some remarks on the case that $\Omega$ is infinite.
In this case, there may not
be a ``closest'' world to $w$ satisfying $\phi$.  As a result, the
definition of $\Cond$ used here based on an ordering
gives counterintuitive results.
It is perhaps
easier to understand this issue in semantic terms, using the
selection function $\ftt_\preceq$.
For example, let
$\Omega^\infty = \{0,1,2,\ldots,\infty\}$ and for all $i$, we have
$i+1 \prec^\infty i$, that is, $i+1$ is closer to $\infty$ than $i$.
Then if $H = \Omega - \{\infty\}$, we would have $\ftt_\preceq(\infty,H)
= \emptyset$ and $\infty \in H \RCond \emptyset$ according to the
definition above, because there is no world ``closest''
to $\infty$ in $H$.
This does not accord with the usual intuitions for $\RCond$.

This particular problem disappears if we require $\preceq^w$ to be
{\em well-founded}, which means that there are no infinitely descending
$\preceq^w$ sequences of the form $\ldots w_n \preceq^w w_{n-1}
\preceq^w \cdots \preceq^w w_0$, as is essentially done by
Kraus, Lehmann, and Magidor \citeyear{KLM}.
Lewis gives a definition that seems to give us the properties we want
even if $\preceq^w$ is not well-founded.
Roughly speaking, his definitions says that $\phi \Cond \psi$ holds at
$w$ if all worlds sufficiently close to $w$ that satisfy $\phi$ also
satisfy $\psi$.
More precisely, Lewis defines $\Cond$ as follows.  Given a  preferential
structure $M = (W,R,\pi)$, we have
\begin{quote}
$(M,w) \sat \phi\Cond\psi$, if  for every
world $w_1 \in \intension{\phi}_M$,
there is a world $w_2$ such that (a) $w_2 \preceq^w w_1$
(so that $w_2$ is at least as close to $w$ as $w_1$),
(b) $w_2 \in \intension{\phi\land\psi}_M$, and (c) for all
    worlds $w_3 \prec^w w_2$, we have
     $w_3 \in \intension{\phi \rimp \psi}_M$ (so any world closer to $w$
than than $w_2$ that satisfies $\phi$ also satisfies $\psi$).
\end{quote}
It is not hard to show that Lewis's definition coincides with that given
here if $\prec$ is well-founded (and, in particular, if $W$ is finite).
Moreover, with this definition, Theorem~\ref{char2syn} holds even if $W$
is infinite (except that P3 has to be strengthened to require that
$\preceq^w$ be well founded as well as linear in order to get it to
correspond to C4).  Thus, it seems that by taking Lewis's definition, we
get precisely the properties we want.

Unfortunately, as results of \cite{FrHK1} show, appearances here are
somewhat deceiving.  Lewis's definition
is still not appropriate for counterfactual or nonmonotonic reasoning
in infinite domains,
once we have a rich enough language.   In \cite{FrHK1}, the language
considered is first-order conditional logic, but the problems can be
demonstrated using the set-theoretic approach as well.

We can easily define an operator
$\RCond$ that captures Lewis's definition, although it does not
correspond to a selection function.  For example,
C0$'$ does not hold with this definition in general (even though it does
hold if $\RCond$ is defined in terms of a selection function).  Consider
the
domain $\Omega^\infty$ above, and let $H_k = \{k, k+1, k+2, \ldots \}$.
Then we have $\infty \in H_0 \RCond H_k$ for all $k$, but since
$\inter_k H_k = \emptyset$, we have $(H_0 \RCond \inter_k H_k) =
\emptyset$.

The fact that C0$'$ does not hold in general is not bad.  It is not
clear that we want it for counterfactual and nonmonotonic
reasoning in the infinite case.
For example, consider a lottery.  If we think of $E_j$ as
corresponding to ``player $j$ wins the lottery'' and $J$ as being the
set of players, then we might well
want to have $\inter_{j \in J} (W \RCond \neg E_j)$, which just says
that, for each player $j$ in $J$, normally
player $j$ does not win the lottery (giving $\RCond$ a normality
reading) and, in addition, $W \RCond (\union_{j \in J} E_j)$, which says
that
normally someone wins the lottery.  But this is incompatible with C0$'$.

While Lewis's definition does not force C0$'$,
results of \cite{FrHK1} show that other properties do hold with
Lewis's definition that are arguably just as undesirable as C0$'$ in the
infinite case.  For example, it is
easy to show that the following property holds,
for any index set $J$:
\begin{equation}\label{eq11}
((\union_{j \in J} H_j) \RCond \neg H_1) \inter \inter_{j \in J}( (H_1
\union H_j) \RCond H_1) = \emptyset.
\end{equation}
(\ref{eq11}) encodes a variant of the lottery paradox.  Consider a
lottery with $J$ players, where player 1 has bought more tickets than
any other player.  It might then seem reasonable to say that
player 1 is more likely to win than any other player, but still unlikely
to win.   If we think of $H_i$ as
corresponding to ``player $i$ wins the lottery'' and we give $\RCond$ a
``typicality'' reading, then this is exactly what the
left-hand side of (\ref{eq11}) says.   However, the fact that the
right-hand side is the empty set says that this situation cannot happen,
according to Lewis's definition.%
\footnote{If $J$ is finite, then (\ref{eq11}) follows easily from C5$'$,
C8$'$, and C10$'$.  It would follow from an infinitary version of C5$'$
if $J$ is infinite, but the infinitary version of C5$'$ does not follow
from the other properties.}

As shown in \cite{FrHK1}, there is another approach that can be used for
giving semantics to conditional logic that involves {\em plausibility
measures} \cite{FrH5Full}, which works appropriately even for
first-order conditional logic (with infinite domains).
Of course, this approach too can be captured by a
set-theoretic approach, but the details of that would take us well
beyond the scope of this paper.

\section{Conclusion}\label{conclusion}
The goal of this paper is to show that we can still get the benefit of
an axiomatic proof theory even if we work at the semantic level.
Indeed, at the semantic level we may get more axioms and easier
completeness proofs.  This should not be interpreted as an argument to
abandon the more traditional, syntactic approach.
Syntactic methods have their place, particularly when we do not have one
fixed model in mind about which we are reasoning.  However, these
results are further evidence showing that when we are working with
a fixed model, semantic reasoning can be a powerful tool.

\paragraph{Acknowledgements:}  I would like to thank Giacomo Bonanno and
the two anonymous referees for their insightful
comments on the paper.

\bibliographystyle{chicago}
\bibliography{z,refs,joe}
\end{document}